\documentclass[lettersize,journal]{IEEEtran}
\usepackage{amsmath,amsfonts}
\usepackage{algorithmic}
\usepackage{algorithm}
\usepackage{array}
\usepackage{graphicx}
\usepackage{caption}    
\usepackage{subcaption} 
\usepackage{textcomp}
\usepackage{stfloats}
\usepackage{url}
\usepackage{verbatim}
\usepackage{graphicx}
\usepackage{cite}
\usepackage{bm}
 \usepackage{amsmath} 
 \usepackage{amssymb} 
\usepackage{gensymb}
\usepackage{xcolor}
\usepackage{hyperref}
\usepackage{multirow}
\usepackage{booktabs}
\makeatletter
\DeclareRobustCommand\onedot{\futurelet\@let@token\@onedot}
\def\@onedot{\ifx\@let@token.\else.\null\fi\xspace}

\makeatother

\definecolor{mygreen}{RGB}{0, 175, 0}
\definecolor{myblue}{RGB}{0, 102, 204}
\definecolor{lightblue}{RGB}{232, 244, 248}
\definecolor{myred}{RGB}{220, 60, 60}
\definecolor{myorange}{RGB}{255, 128, 0}
\definecolor{greenyellow}{RGB}{197,228,10}

\begin{document}

\title{Generic Objects as Pose Probes for Few-shot View Synthesis}

\author{Zhirui Gao, Renjiao Yi\textsuperscript{$\dagger$} , Chenyang Zhu, Ke Zhuang, Wei Chen, Kai Xu\textsuperscript{$\dagger$}, ~\IEEEmembership{Senior Member,~IEEE}

\thanks{This work is supported in part by the NSFC (62325211, 62132021, 62372457), the Major Program of Xiangjiang Laboratory (23XJ01009), Young Elite Scientists Sponsorship Program by CAST (2023QNRC001), the Natural Science Foundation of Hunan Province of China (2022RC1104) and the NUDT Research Grants (ZK22-52).}

\thanks{All authors are with the School of Computer, National University of Defense Technology, Changsha, 410073, China.}
\thanks{ {$\dagger$} Corresponding authors: Renjiao Yi and Kai Xu (E-mails: yirenjiao@nudt.edu.cn, kevin.kai.xu@gmail.com).}

}



\markboth{IEEE Transactions on Circuits and Systems for Video Technology}
{Shell \MakeLowercase{\textit{et al.}}: A Sample Article Using IEEEtran.cls for IEEE Journals}

\maketitle

\begin{abstract}
Radiance fields, including NeRFs and 3D Gaussians, demonstrate great potential in high-fidelity rendering and scene reconstruction, while they require a substantial number of posed images as input. COLMAP is frequently employed for preprocessing to estimate poses. However, COLMAP necessitates a large number of feature matches to operate effectively, and struggles with scenes characterized by sparse features, large baselines, or few-view images. 
We aim to tackle few-view NeRF reconstruction using only 3 to 6 unposed scene images, freeing from COLMAP initializations. 
Inspired by the idea of calibration boards in traditional pose calibration, we propose a novel approach of utilizing everyday objects, commonly found in both images and real life, as ``pose probes''.  
By initializing the probe object as a cube shape, we apply a dual-branch volume rendering optimization (object NeRF and scene NeRF) to constrain the pose optimization and jointly refine the geometry.  PnP matching is used to initialize poses between images incrementally, where only a few feature matches are enough. 
PoseProbe achieves state-of-the-art performance in pose estimation and novel view synthesis across multiple datasets in experiments. We demonstrate its effectiveness, particularly in few-view and large-baseline scenes where COLMAP struggles. In ablations, using different objects in a scene yields comparable performance, showing that PoseProbe is robust to the choice of probe objects.  Our project page is available at: \href{https://zhirui-gao.github.io/PoseProbe.github.io/}{https://zhirui-gao.github.io/PoseProbe.github.io/}
\end{abstract}


\IEEEpubid{\begin{minipage}{\textwidth}\ \centering
    Copyright \copyright 2025 IEEE. Personal use of this material is permitted. \\However, permission to use this material for any other purposes must be obtained from the IEEE by sending an email to pubs-permissions@ieee.org. 
\end{minipage}}

\IEEEpubidadjcol

\begin{figure}[t]
\centering
\includegraphics[width=\linewidth]{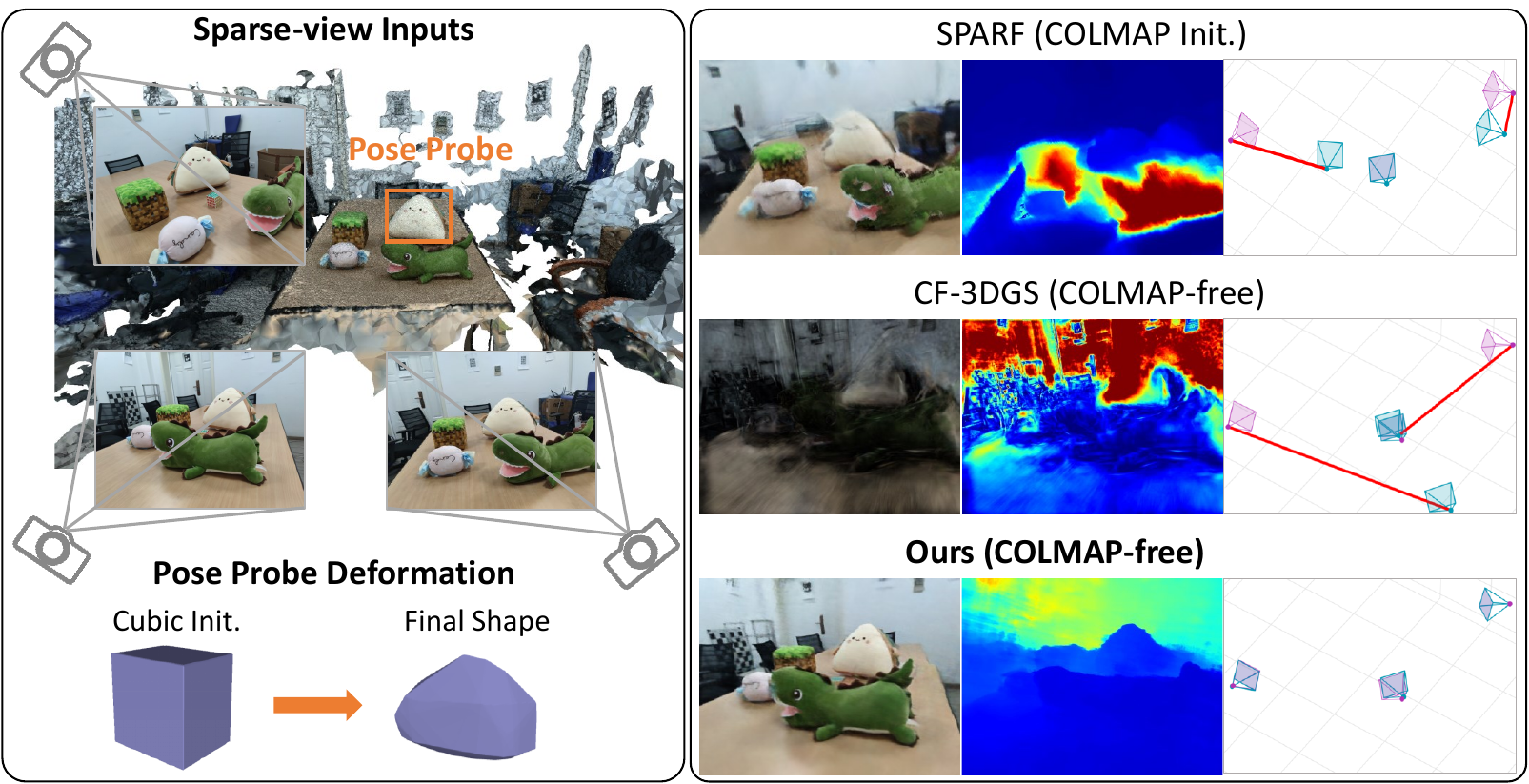}
\caption{Our method addresses pose estimation and NeRF-based reconstruction in the challenging few-view setting (only 3 unposed views).  
Most NeRF-based approaches are initialized from COLMAP poses. However, in sparse regimes, COLMAP often fails to initialize, making it challenging for pose optimization in a state-of-the-art method SPARF~\cite{truong2023sparf} to work well.  A state-of-the-art COLMAP-free pipeline CF-3DGS~\cite{Fu_2024_CVPR} also struggles in sparse-view scenarios.  We propose spotting generic objects as ``pose probes'' in the scene 
(a face-shaped toy in this example), it achieves realistic novel-view renderings and accurately reconstructs geometry using only 3 input images.
}
\label{fig:teaser}
\end{figure}
\section{Introduction}\label{intro}

\IEEEPARstart{A}s a milestone in the realm of computer vision and graphics, neural radiance fields (NeRFs) offer unprecedented capability of photorealistic rendering of scenes from multi-view posed images. The accuracy of novel-view renderings depends heavily on the precision of input camera poses and the number of input images, limiting its applicability in real-world scenarios. Camera poses of the input views are typically recovered with COLMAP~\cite{7780814} in most prior works. However, we find that if input images are too few and views are sparse, COLMAP often fails to work due to wide baselines and insufficient feature matches.

To alleviate reliance on accurate input poses, many studies estimate or refine them based on various assumptions. Earlier works~\cite{wang2021nerfmm,lin2021barf, meng2021gnerf, chen2023local, jeong2021self} predominantly use photometric losses to optimize NeRFs and camera poses jointly. For example, NeRFmm~\cite{wang2021nerfmm} focuses on forward-facing scenes where the baseline is relatively small. BARF~\cite{lin2021barf} proposes a coarse-to-fine frequency encoding strategy to facilitate joint optimization. GNeRF~\cite{meng2021gnerf} on the other hand, relies on pose distribution assumptions. However, in sparse-view scenarios, photometric losses alone often prove insufficient for accurate pose estimation due to the under-constrained nature of 3D reconstruction. To address this limitation, methods such as Nope-NeRF~\cite{bian2023nope} and CF-3DGS~\cite{Fu_2024_CVPR} leverage monocular depth estimation from dense video frames to introduce additional constraints. However, their reliance on dense input frames significantly restricts their applicability in few-view cases. More recently, SPARF~\cite{truong2023sparf}  and TrackNeRF~\cite{truong2023sparf} have advanced pose optimization for few-view settings by utilizing reprojection consistency, marking notable progress in pose-optimized NeRF development.  Despite these advances, such methods still require reasonable pose initialization and cannot operate entirely without prior pose information.  \IEEEpubidadjcol
Consequently, reconstructing NeRFs in few-view scenarios without any pose priors remains a significant challenge in the field. In this paper, we propose a COLMAP-free approach tailored for few-view inputs, as demonstrated in Fig.~\ref{fig:teaser}.

A traditional way for obtaining accurate poses involves placing a calibration board in the scene. However, such boards are rarely present in real-world, everyday environments. This limitation motivates us to explore the potential of leveraging in-situ everyday objects as natural calibration probes. These objects, already present in the scene, offer a practical and low-effort alternative to artificial calibration tools. Upon spotting a probe object, we employ Grounded-SAM~\cite{ren2024grounded} to automatically segment it using text prompts, followed by shape initialization with a cuboid. The core innovation of our approach lies in repurposing existing scene objects as calibration probes, enabling efficient and accurate camera pose estimation. We focus on exploring two main benefits of pose probes: (1) An incremental camera pose optimization strategy addresses the challenge of missing initial poses. New views are added incrementally, with initial poses estimated using the Perspective-n-Point, which leverages the initial probe’s geometry and image correspondences. (2) Pose probes establish strong pose constraints, which significantly enhance pose accuracy and, in turn, improve the quality of novel view rendering.   We find that most everyday objects
can be efficiently employed as pose probes (Fig.~\ref{fig:eff_probe}), the method is insensitive to the choice of probe objects. 

\begin{figure*}[t]
\centering%
\includegraphics[width=0.95\textwidth]{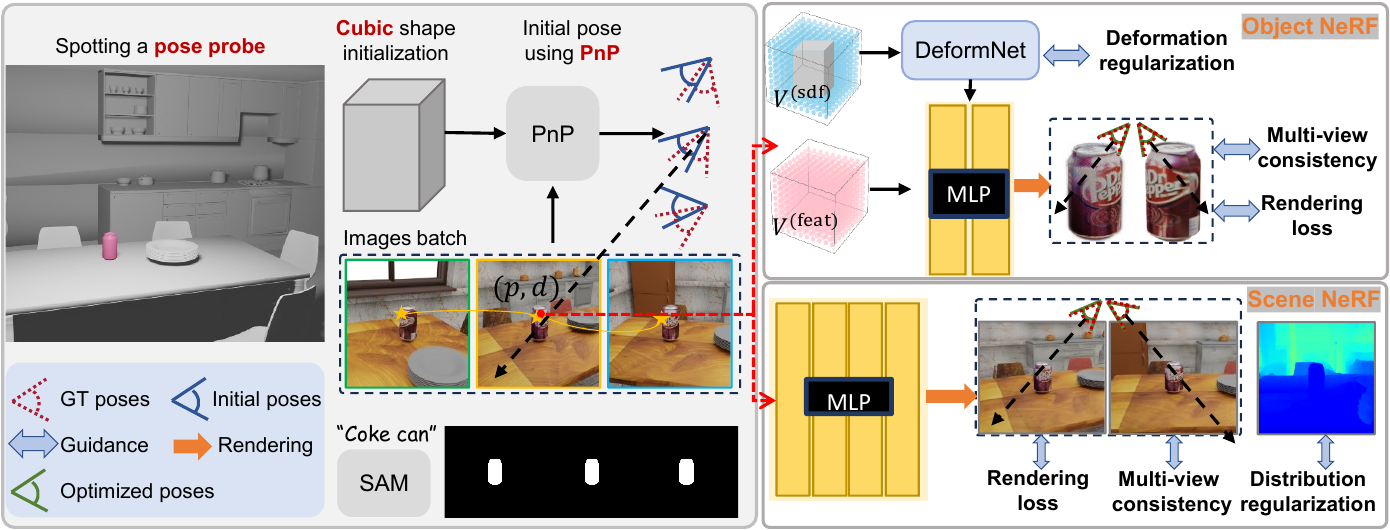}
\caption{Method overview. Our approach utilizes generic objects as pose probes for few-view inputs, with their masks automatically segmented using Grounded-SAM~\cite{ren2024grounded} with text prompts. The pose probe is initialized as a cuboid and then used to estimate the initial poses of input images via PnP incrementally. Our pipeline employs a dual-branch volume rendering framework to optimize camera poses and scene representation jointly. In the object NeRF branch, a hybrid SDF representation models the object geometry, enforcing constraints such as deformation regularization, multi-view consistency, and rendering loss. The scene branch optimizes the entire scene within an implicit radiance field. We refine camera poses simultaneously, incorporating constraints like rendering loss, multi-view consistency, and distribution regularization, yielding precise pose estimation.}

\label{fig:pip}
\end{figure*}

We introduce a pipeline for NeRF reconstruction from few-view (3 to 6) unposed images.  The overall pipeline is shown in Fig.~\ref{fig:pip}, where a dual-branch volume rendering optimization workflow is adopted, targeting the probe object and the entire scene respectively. 
The object branch employs a hybrid architecture designed for efficient volume rendering. Geometry is represented using a signed distance field (SDF) voxel grid, serving as a template field, combined with an implicit DeformNet to capture deformations. The SDF voxel grid is initialized as a cuboid, representing the initial shape of the pose probe, which is subsequently refined through deformation by the DeformNet.  Appearance is represented by a feature grid, and all components are unified through a shallow MLP to enable volume rendering, facilitating the joint optimization of camera poses and object geometry. To enforce learning accurate camera poses and object geometry, the training process incorporates constraints such as multi-view geometric consistency and deformation regularization.  Meanwhile, the scene branch is designed to learn a neural representation of the entire scene while refining camera poses. To further enhance convergence in both pose estimation and scene reconstruction, multi-layer feature-metric consistency is introduced as an additional constraint.

The initial camera poses for the first two views are estimated using PnP, leveraging the correspondences of the pose probe between two images and its initial cuboid shape.  Subsequent frames are incrementally incorporated during the optimization, using PnP to estimate initial poses. Poses are further optimized jointly by both branches to get final results.  Note that PnP matching requires only several feature matches, working for feature-sparse scenarios, while COLMAP often fails due to insufficient feature matches. As tested in Tab.~\ref{tab:pose_init}, even using an identity matrix or very noisy poses (adding 30\% noises to PnP poses), the method still gets comparable performance with a slight drop in metrics.  In this way, we obtain high-quality novel view synthesis and poses, without any pose priors, even for large-baseline and few-view images. As shown in Fig.~\ref{fig:teaser}, aided by the proposed pose probe (the face toy), our method produces realistic novel-view renderings and accurate poses using only three input images, without relying on pose initialization, outperforming both COLMAP-based and COLMAP-free state-of-the-art methods.

The main contributions include:
\begin{itemize}
    \item We utilize generic objects as pose calibration probes, to tackle the challenging few-view and feature-sparse scenes using only 3 to 6 images, where COLMAP may be inapplicable. 
    \item We propose a dual-branch approach targeting both the object and the scene, where we introduce multi-view geometric consistency and multi-layer feature-metric consistency as novel training constraints within a joint pose-NeRF training pipeline.
   
    \item We introduce a texture-sparse scene dataset, and compare the proposed method with SOTAs across four benchmarks, where our method achieves PSNR improvements of 14.3\%, 5.09\%, 1.93\%  and 5.30\% in novel view synthesis on average, along with significant enhancements in pose metrics. The proposed method successfully handles sparse-view scenes where COLMAP experiences a 67\% initialization failure rate. 
\end{itemize}

\section{Related Works}\label{relatedwork}

\subsection{Radiance fields with pose optimization} The reliance on high-precision camera poses as input restricts the applicability of NeRFs and 3D Gaussian Splatting~\cite{kerbl20233d} (3DGS). Several studies have sought to alleviate this dependency. NeRF-based techniques utilize neural networks to represent the radiance fields and jointly optimize camera parameters, as demonstrated by early approaches ~\cite{wang2021nerfmm, jeong2021self,lin2021barf,chng2022garf}. NeRFmm~\cite{wang2021nerfmm} jointly optimizes the radiance field and camera parameters and works on forward-facing scenes. BARF~\cite{lin2021barf} introduces a coarse-to-fine positional encoding strategy, making the joint pose and scene optimization easier. L2G-NeRF~\cite{chen2023local} integrated a local-to-global registration into BARF to improve anti-noise ability. SCNeRF~\cite{jeong2021self}  further optimizes camera distortion and proposes a geometric loss to regularise learned parameters. GARF~\cite{chng2022garf} and SiNeRF~\cite{xia2022sinerf} employ different activation functions to overcome local
minima in optimization~\cite{liu2022efficient}. LU-NeRF~\cite{cheng2023lu} operates in a local-to-global manner to achieve a final global optimization of camera poses and scenes without pose assumption. CRAYM~\cite{CRAYM24} utilizes the correspondence of camera rays to optimize pose and scene representation jointly. This approach not only produces more realistic renderings but also reconstructs finer geometric details. Additionally, Camp~\cite{10.1145/3618321} proposes using a proxy problem to compute a whitening transform, which helps refine the initial camera poses. 
Based on RGB-D video sequences, visual SLAM approaches with neural scene representations~\cite{rosinol2022nerf,zhu2022nice} significantly improve the scene and pose estimation processes. For instance,  NoPe-NeRF~\cite{bian2023nope} adopts monocular depth estimation to learn scene representation without depending on pose priors. GC-NeRF~\cite{zhang2024learning} establishes the photometric constraints and s explores geometry projection during training. SPARF~\cite{truong2023sparf} and TrackNeRF~\cite{10.1007/978-3-031-73254-6_27} address the challenge of NeRFs with sparse-view, wide-baseline input images but require initial camera positions near ground truth, which limits its applicability in real-world scenarios.

3DGS-based methods employ explicit 3D Gaussians instead of neural networks and have been studied extensively. CF-3DGS~\cite{Fu_2024_CVPR} and COGS~\cite{COGS2024} use monocular depth estimators to aid camera pose registration. Recent work~\cite{fan2024instantsplat} uses an off-the-shelf model~\cite{dust3r_cvpr24} to compute initial camera poses for sparse-view, SfM-free optimization. However, 3DGS relies on an initial point cloud, which is hard to acquire in unconstrained scenes with few views and unknown poses. Consequently, 3DGS-based methods generally rely on pre-trained vision models~\cite{dust3r_cvpr24, Ranftl2020}, significantly increasing complexity.

\subsection{Novel-view synthesis from few views}  
Various regularisation techniques shine in few-view learning to address the challenge of requiring dense input views. Some work~\cite{kangle2021dsnerf, guo2024depth,10510490} utilize depth supervision to avoid overfitting and facilitate better geometry learning. 
Additionally, appearance regularization~\cite{Niemeyer2021Regnerf}, geometry regularization~\cite{song2022neural,Niemeyer2021Regnerf,liu2024georgs} and frequency regularization~\cite{yang2023freenerf, truong2023sparf} are introduced to optimize the radiance fields. In addition, AR-NeRF~\cite{xu2025few} further investigates the inconsistency between the frequency regularization of Positional Encoding and rendering loss, and proposes an adaptive rendering loss regularization to align the frequency relations between them. Some recent works~\cite{zhu2023FSGS, xu2024mvpgs, li2024dngaussian, yin2024fewviewgs, peng2024neurips} combine the efficient 3DGS representation and additional geometric constraints(monocular depth estimation~\cite{ranftl2021vision} or correspondences consistency) to improve the efficiency of novel-view synthesis in sparse-view scenarios.

\subsection{Object pose estimation} 
Object pose estimation is a long-studied problem closely related to camera pose optimization. A popular class of pose estimation methods~\cite{shugurov2021dpodv2,rothganger20063d,peng2022pvnet,zhong2022sim2real} establishes dense 2D-3D correspondences between image pixels and the surface of a 3D object. These matching-based approaches typically use PnP+RANSAC~\cite{lepetit2009ep, fischler1981random} to compute the 6D object pose. In this paper, we adopt this paradigm to estimate the initial camera poses. However, since the exact object geometry is unknown a priori, we propose to optimize camera poses incremental and object geometry within a NeRF-based pipeline. CAD model-free pose estimation methods~\cite{sun2022onepose,li2023nerf} reconstruct the object beforehand, but usually lack good generalization and are difficult to meet the high-precision requirements of NeRF. Other works~\cite{zhang2022relpose, lin2023relpose++} demonstrate the capability to recover 6D poses from sparse-view observations, but these poses still require refinement during NeRF training.

\section{Method}\label{method}
We propose a dual-branch pipeline as illustrated in Fig.~\ref{fig:pip}, which combines both neural explicit and implicit volume rendering. The essential preliminaries are presented in Sec.\ref{subsec:preliminaries}. To effectively leverage the advantages of the pose probe, we design a neural volume rendering with an explicit-implicit signed distance field (SDF) in the object branch (Sec.\ref{subsec:stage_0}). The scene branch (Sec.\ref{subsec:stage_1}) employs an implicit NeRF to achieve a high-quality representation while optimizing the camera pose. Finally, joint training is described in Sec.~\ref{subsec:overall}.

\subsection{Preliminaries}\label{subsec:preliminaries}
\textbf{Explicit volumetric representation.} DVGO~\cite{sun2022direct}  optimizes a density voxel grid $\bm{V}^{\text{(density)}} \in \mathbb{R}^{1 \times N_x \times N_y\times N_z}$ and a feature voxel gird $\bm{V}^{\text{(feat)}} \in \mathbb{R}^{D \times N_x \times N_y\times N_z}$  with a shallow MLP parameterized by $\Theta$, where $D$ is a hyperparameter for feature-space dimension. Given a 3D position $\bm{p}$ and viewing-direction $\bm{d}$, the volume density $\bm{\sigma}$ is estimated by:  
\begin{equation}
 \bm{\sigma} = \operatorname{interp}\left(\bm{p},\bm{V}^{\text{(density)}}\right),
\end{equation}
where ``interp'' denotes trilinear interpolation. The color $\bm{c}$ is performed by:
\begin{equation}
\bm{c} = \operatorname{MLP}_{\Theta}\left(\operatorname{interp}(\bm{p}, \bm{V}^{\text{(feat)}}), \bm{p}, \bm{d}\right). \label{eq:queries_color}
\end{equation}
The positional encoding~\cite{mildenhall2020nerf} is applied for both $\bm{p}$ and $\bm{d}$. 

\textbf{Neural SDF representation.}
With neural volume rendering, the estimated colors of  the image can be presented as:
\begin{equation}
\hat{C}=\sum_{i=1}^n{w\left( t_i \right) \hat{c}\left( t_i \right)},
\end{equation}
where $t_i$ is the depth of the $i$ th of point sampled along the ray, $w\left( t_i \right)$ and $\hat{c}\left( t_i \right)$ is the weight and the estimated color field for the point at $t_i$.
\begin{equation}
w\left( t_i \right) = T\left( t_i \right) \alpha\left(t_i \right), T\left( t_i \right) = \sum_{j=1}^{i-1}(1-\alpha\left(t_j \right)),
\end{equation}
where $T\left( t_i \right)$ is the accumulated transmittance and  $\alpha\left(t_i \right)$ is the opacity value. 

\subsection{Object NeRF with pose estimation}\label{subsec:stage_0}

The object NeRF aims to provide initial camera poses and establish camera pose constraints by modeling the pose probe. Inspired by the fast convergence of explicit representations~\cite{sun2022direct,wu2022voxurf} while maintaining high-quality rendering, we design a similar volume rendering framework for the object branch. To recover high-fidelity geometry and initial camera poses efficiently, we employ an explicit-implicit SDF representation~\cite{wang2021neus,fu2022geo} for the object NeRF. To better leverage object geometry, the gradient at each point is embedded into the color rendering process: 
\begin{equation}
\bm{c} = \operatorname{MLP}_{\Theta}\left(\operatorname{interp}(\bm{p}, \bm{V}^{\text{(feat)}}), \bm{n}, \bm{p}, \bm{d}\right) .
\end{equation}
Here, the normal $\bm{n}$ is computed as the normalized gradient of the SDF, and $\bm{d}$ represents the viewing direction.   

In this section, we delve into the hybrid explicit-implicit representation of the object branch. Following this, we present our incremental pose optimization strategy, which computes the initial camera poses for each frame and further refines them. Finally, we describe how geometric consistency is leveraged to jointly optimize the neural fields and camera poses, using the constraints provided by the pose probe for robust and accurate reconstruction.

\begin{figure}[t!]
\centering
\includegraphics[width=0.5\textwidth]{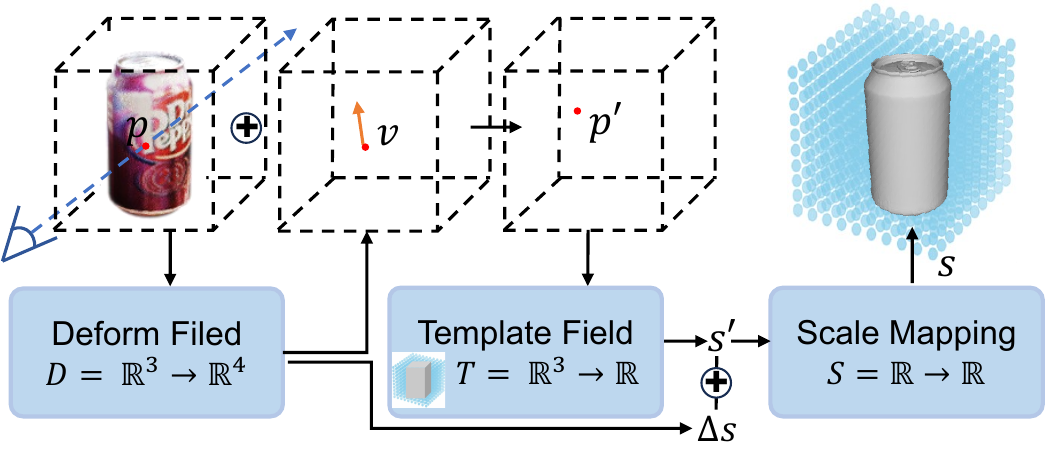} \\
\caption{
Overview of the hybrid SDF representation. For a given point $\bm{p}$, the implicit deformation field predicts a deformation vector $\bm{v}$ and a scalar correction $\Delta s$. The point position is first deformed to $\bm{p}^{\prime} = \bm{p} + \bm{v}$, and  $s'$ is queried from the template field. The final SDF value $s$ is computed by a non-linear scale mapping function $S$ to the sum of $s'$ and $\Delta s$.  
}
\label{fig:deformnet}

\end{figure}

\subsubsection{Hybrid SDF representation} 
The design of our explicit-implicit SDF generation network leverages the initialized cuboid as a geometric prior, which has been shown to be critical for reconstruction accuracy and efficiency~\cite{zhang2021ners, deng2021deformed}. The explicit template field $T$ is a non-learnable voxel grid $\bm{V}^{\text{(sdf)}}$, while the implicit deformation field $D$ is implemented as an MLPs to predict a deformation field and a correction field on top of $T$, as illustrated in Fig.~\ref{fig:deformnet}. The voxel grid $\bm{V}^{\text{(sdf)}}$ is initialized with a coarse template (e.g., a cuboid ), which we find sufficient for learning detailed geometry and enforcing camera pose constraints.  
The template field $T$ serves as a strong prior, significantly reducing the search space and enabling detailed geometry reconstruction with fewer parameters. We provide an ablation study in Sec.~\ref{sec:comp_abl} demonstrating that initializing with a template field leads to faster convergence and more accurate geometry compared to learning the SDF field from scratch.  

To capture shape details, we employ an implicit deformation field $D$ to refine the coarse SDF. Although optimizing an explicit SDF voxel grid on top of the template field $T$ appears more straightforward, it frequently leads to degenerate solutions and geometric artifacts in sparse-view inputs, as discussed in Sec.~\ref{sec:obj_rec}. In contrast, the implicit field inherently provides a smooth and continuous representation, which is well-suited for capturing fine details and complex deformations. Inspired by~\cite{deng2021deformed}, Our deformation field  $D$ predicts a deformation vector $v$ and a scalar correction value $\Delta s$ for each point $\bm p$:
\begin{equation}
D : \bm{p} \in \mathbb{R}^{3} \longrightarrow (v, \Delta s) \in \mathbb{R}^{4}
\end{equation}

The ultimate shape
is determined by interpolating at its deformed location within the template field $T$, further refined by a correction scalar. Therefore, the SDF value of the point $\bm p$ is represented as:  

\begin{equation}
\text{SDF}(\bm{p}) = T(\bm{p} + D_{v}(\bm{p})) + D_{\Delta s}(\bm{p}). 
\label{eq:qurey_sdf}
\end{equation}

\begin{figure}[tb]
\centering
\includegraphics [width=0.48\textwidth]{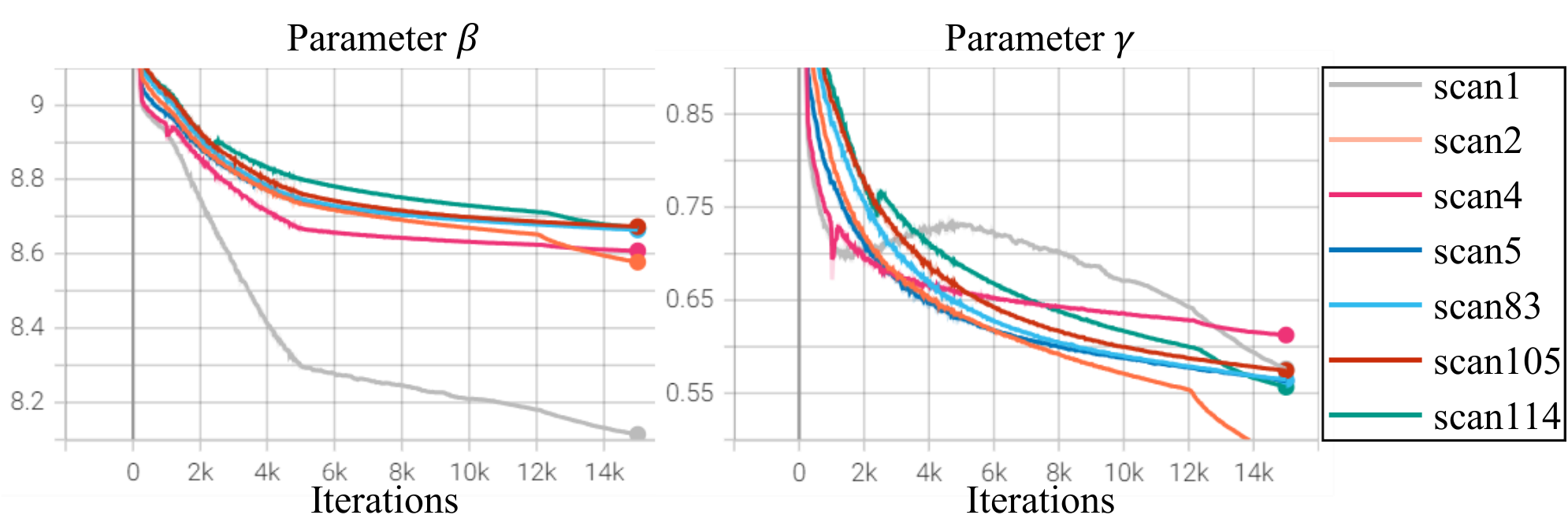} \\
\caption{ 
The adaptive tuning of the training parameters  $\beta$ and $\gamma$ in the SDF mapping function is demonstrated across different scenes in the DTU dataset. The parameters vary from scene to scene, with each scene represented by a distinct colored curve.}
\label{fig:alpha-beta}
\end{figure}

Although the $\text{SDF}(\bm{p})$ is utilized to estimate the volume opacity, directly employing it in Eqn.~\ref{eq:qurey_sdf} is suboptimal for volume rendering due to the manually predefined value scale. To this end, we propose a mapping function $S$ with two learnable parameters to scale the original SDF to the scene-customized scale:
\begin{equation}
{S (\bm{p}})  = \beta (1/(1+e^{-\gamma \cdot SDF (\bm{p})}) -0.5), \label{eq:mapping}
\end{equation}

\noindent where $\beta$ and $\gamma$ are trainable parameters to control the scale of $\bm{V}^{\text{(sdf)}}$. To maintain the original SDF sign, we apply the Softplus activation function to $\beta$ and $\gamma$. The $\beta$ and $\gamma$ vary from scene to scene as illustrated in Fig~\ref{fig:alpha-beta}. The hybrid SDF merges the advantages of explicit and implicit representations, balancing rapid convergence with detailed modeling.

\subsubsection{Incremental pose optimization}  
Our method employs an incremental pose optimization framework to improve initial camera poses and reduce optimization complexity. This framework progressively incorporates new camera views into the training loop, assuming known camera intrinsic parameters $K$. The process begins by initializing the reference image $I_0$ through multi-view mask alignment: we perturb the initial camera orientation to sample candidate projection views, render object masks using the intrinsics $K$, and select the pose $\hat{P}_0$ with the highest mask similarity as the initial pose of the first view. For each subsequent image $I_{i+1}$,  2D correspondences with the previous frame $I$ are first extracted using SuperGlue~\cite{sarlin2020superglue} with SuperPoint~\cite{detone2018superpoint}. These matched 2D points in $I_{i}$ are projected onto 3D surface positions using the optimized pose $\hat{P}_i$ and intrinsics $K$, establishing 2D-3D correspondences for $I_{i+1}$, as elaborated in the next paragraph. Then, the initial pose is robustly estimated using PnP-RANSAC, which simultaneously minimizes reprojection errors and removes outliers. Finally, the new pose $\hat{P}_{i+1}$, existing poses, and radiance field are jointly optimized using photometric and geometric consistency constraints.

 With the explicit SDF voxel grid with shallow DeformNet, the surface points corresponding to image pixels can be efficiently estimated through ray casting, similar to~\cite{fu2022geo,oechsle2021unisurf}. 
Given a camera pose ${\hat{P}}$, intrinsic parameters $K$, and a pixel coordinate  $\mathbf{x} \in \mathbb{R}^2$, we can compute the corresponding ray in the world coordinate system. The 3D sampled points along this ray are expressed as $\bm{p}_i=\bm{o} + t_i\bm{v}$, where $t_i$ represents the $i$-th sampled depth. To identify an occlusion-aware surface point, we find the smallest index $i$ such that the sign of the SDF value at $\bm{p}_{i}$  and $\bm{p}_{i+1}$ differ:
\begin{equation}
i^* = \text{argmin} \left\{i|S(\bm{p}_i) \cdot S(\bm{p}_{i+1})<0 \right\}.\label{eq:find_t}
\end{equation}
Linear interpolation is applied along the  line $\bm{p}_i\bm{p}_{i+1}$ to compute the surface point $\mathbf{S}$ associated with $x$ and camera pose $\hat{P}$:

\begin{equation}
\mathbf{S}(\hat{P},x) =  \bm{o}+ \hat{t}\bm{v}, 
\hat{t}=\frac{S(\bm{p}_{i^*}) \bm{p}_{i^*+1}-{S}(\bm{p}_{i^*+1}) \bm{p}_{i^*}}{{S}(\bm{p}_{i^*})-{S}(\bm{p}_{i^*+1})} . \label{eq:inter_surface}
\end{equation}

\subsubsection{Geometric consistency} 
\begin{figure}[t!]
\centering
\includegraphics[width=0.5\textwidth]{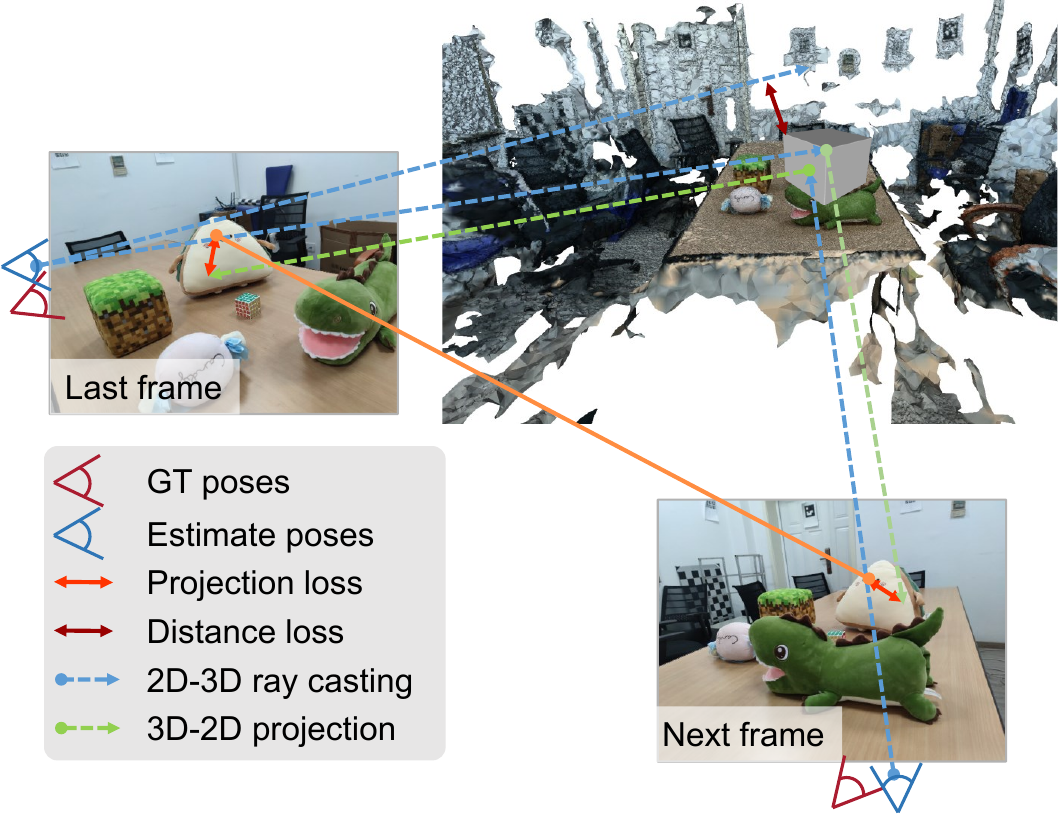} \\
\caption{Illustration of multi-view geometric consistency and ray distance loss. Multi-view geometric consistency ensures accurate alignment of corresponding points across multiple views by minimizing the reprojection error, while the ray distance loss regularizes the minimal distance between camera rays and the surface of the pose probe. Together, these contribute to improved scene reconstruction and camera pose estimation.
}
\label{fig:geo_loss}

\end{figure}

Previous studies~\cite{truong2023sparf, bian2023nope, 10.1007/978-3-031-73254-6_27} have demonstrated that photometric loss alone is insufficient for achieving globally consistent geometry in pose/NeRF systems. To tackle this, we investigate geometric consistency constraints aided by the pose probe. Drawing inspiration from SPARF~\cite{truong2023sparf} and TrackNeRF~\cite{10.1007/978-3-031-73254-6_27}, which use reprojection error to ensure consistency, we adopt a more direct multi-view projection distance to constrain camera poses, as illustrated in Fig.~\ref{fig:geo_loss}. Formally, for an image pair $\left( I_i, I_j \right)$ and matching pixel pairs $\left(\mathbf{x}, \mathbf{y}\right)$,  we first locate the surface points $\left(\mathbf{S_x}, \mathbf{S_y}\right)$ using ray-casting. These 3D surface points are then projected back onto the image plane to minimize the distance between correspondences. The geometric circle projection distance of pair $\left(\mathbf{x}, \mathbf{y}\right)$  is defined as:
\begin{equation}
\label{eq:geo-dis}
\mathcal{D}{(\mathbf{x},\mathbf{y})}  =  \rho (  \,  \pi ( \mathbf{S_x}, \hat{P}_j)-\mathbf{y}  )  +  \rho  (\pi ( \mathbf{S_y}, \hat{P}_i)-\mathbf{x}),
\end{equation}
where $\pi$ means the camera projection function, and $\rho$ denotes Huber loss function. Furthermore, based on the prior that rays emitted from feature points should intersect the object, we introduce a regularization term that minimizes the distance between these rays and the surface of the pose probe to refine the camera poses:

\begin{equation}
\label{eq:inter}
\mathcal{L}_{\text{dis}}(\bm{r}, \bm{o}) = \max(\text{dis}(\bm{r}, \bm{o}) - L, 0),
\end{equation}
where $\text{dis}(\bm{r}, \bm{o})$ denotes the shortest distance from the object center $\bm{o}$ to the ray $\bm{r}$, and $L$ indicates the maximum radius of the object. Finally, our multi-view geometric consistency objective is formulated as: 
\begin{equation}
\label{eq:geo-cons}
\mathcal{L}_{\text{GeoObj}}(\hat{\mathcal{P}}) = \sum_{(\mathbf{x},\mathbf{y}) \in \mathcal{V}}  w_{\mathbf{x}}\mathcal{D}{(\mathbf{x},\mathbf{y})} + \lambda \mathcal{L}_{\text{dist}}(\bm{r_{x,y}}, \bm{c}) . 
\end{equation}  
Here, $w_{\mathbf{x}}$ represents the matching confidence associated with pair $\left(\mathbf{x}, \mathbf{y}\right)$ and $\lambda$ is set to 10.

\subsection{Scene NeRF with pose refinement}\label{subsec:stage_1}

While training the object NeRF, we simultaneously optimize a scene NeRF branch to model the entire scene and utilize its global context for joint camera pose refinement. Our scene NeRF builds upon a coarse-to-fine positional encoding baseline~\cite{lin2021barf}, enhanced with the projection distance loss (Eqn~\ref{eq:geo-dis}) to enforce multi-view geometric consistency. To address challenges in joint pose-scene optimization (e.g., geometric mismatches and background distortion), we introduce two additional constraints: multi-layer feature-metric consistency and compact distribution regularization.

\subsubsection{Multi-layer feature-metric consistency} Although geometric consistency facilitates rapid convergence in camera pose optimization, mismatches may introduce incorrect supervisory signals, potentially leading to local optimal. Inspired by dense bundle adjustment~\cite{tang2018ba}, we introduce a multi-layer feature-metric consistency. This constraint minimizes the feature difference of aligned pixels using cosine similarity. The multi-layer feature-metric associated with pixel $\mathbf{x}$ is formulated as:
\begin{equation}
 e_{\mathbf{x}} =  \sum_{k=1}^{M}{ 1- \text{cos} \left( F_{j,k}(\pi(\mathbf{S_x}, \hat{P}_{j})), F_{i,k}(\mathbf{x})\right)}, \label{eq:fea-con}
\end{equation}
where $\mathbb{F} = \{ F_{i,k} | i = 1...N, k = 1...M\}$ are the multi-layer image features extracted by the pretrained VGG~\cite{simonyan2015very}. Here, $N$ denotes the number of images, and $M$ is the number of layers. Our feature-metric loss is defined as: $$\mathcal{L}_{\text{Fea}}(\hat{\mathcal{P}}) =\sum_{x \in \mathcal{V}}  \gamma_{\mathbf{x}} e_{\mathbf{x}}.$$ 

A visible mask $\gamma \in [0,1]$ is applied to filter out out-of-view or occluded points from alternative perspectives. Points whose projected pixels fall outside the image boundary are considered out-of-view, while points with invalid depth values are treated as occluded. To identify occluded pixels, we follow the strategy of SPARF~\cite{truong2023sparf}, which calculates the cumulative density along a ray to detect occupied regions from other views. This constraint considers more image pixels rather than focusing solely on keypoints in the geometric consistency. In contrast to photometric error, which is sensitive to initialization and increases non-convexity~\cite{engel2017direct}, our feature-based consistency loss provides smoother optimization.

\subsubsection{Compact distribution regularization}
We have observed that "background collapse" tends to occur in the training, especially in scenes with few textures. Background collapse is a phenomenon in which distant surfaces tend to gather near the camera~\cite{barron2022mipnerf360}. Inspired by Mip-NeRF 360~\cite{barron2022mipnerf360},  we adopt a similar regularize to encourage the distribution of density in the radiance field to be compact:
\begin{align}
\begin{split}
\mathcal{L_{\text{Dist}}}(\hat{d}, \mathbf{w}) =
\sum_{i,j} w_{i} w_{j} & \left| \frac{\hat{d}_{i} +\hat{d}_{i+1}}{2} - \frac{\hat{d}_{j} + \hat{d}_{j+1}}{2} \right|  \\
&+ \frac{1}{3}\sum _{i} w_{i}^{2}( \hat{d}_{i+1} - \hat{d}_{i}),
\label{eq:dist_loss} 
\end{split}
\end{align}
where $\hat{d}_{i}$ is the sampling depth of $i$-th sampling points and $w_i$ is the alpha compositing.

\subsection{Joint training}\label{subsec:overall}
The final training objectives consist of all losses for the object NeRF and scene NeRF: $\mathcal{L} = \lambda \mathcal{L}_{\text{Obj}} +  \mathcal{L}_\text{Sce}$. 

\subsubsection{Object NeRF}  To encourage smoother
deformation and prevent large shape distortion, we incorporate a deformation regularization on the deformation field and a minimal correction prior~\cite{deng2021deformed} for the correction field:
\begin{equation}
\mathcal{L_\text{d}} = \sum_{\bm{p} \in \Omega} \sum_{d\in{X,Y,Z}} \left \| \nabla D_{v} \small|_{d}  (\bm{p})   \right\|_{2} +  \sum_{\bm{p} \in \Omega} \left|D_{\Delta s} (\bm{p})  \right | .
\label{eq:deform_reg} 
\end{equation}
Besides, we add an Eikonal term~\cite{gropp2020implicit} to   regularize the the SDF:
\begin{equation}
\mathcal{L_\text{r}} =  \sum_{\bm{p} \in \Omega} 
\small| \left \| \nabla S(\bm{p}) \right\|_{2} -1 \small| .
\end{equation}
\begin{equation}
\mathcal{L_\text{Obj}} = \mathcal{L_\text{rgb}} + \mathcal{L_\text{m}} +  \lambda_1 \mathcal{L_\text{GeoObj}}  +  \lambda_2  \mathcal{L_\text{d}} + \lambda_3  \mathcal{L_\text{r}},
\end{equation}
where $\mathcal{L_\text{rgb}}$  and $\mathcal{L_\text{m}}$ represent photometric $l_2$ loss and mask $l_1$ loss respectively. We set $\lambda_1$, $\lambda_2$, and $\lambda_3$ to 0.01, 0.001 and 0.01.

\subsubsection{Scene NeRF} In the scene training stage, the total loss is: 
\begin{equation}
\mathcal{L_\text{Sce}} = \mathcal{L_\text{rgb}} +  \lambda_4 \mathcal{L_\text{GeoSce}} +\lambda_5 \mathcal{L_\text{Fea}} + \lambda_6 \mathcal{L_\text{Dist}},
\end{equation}
where $\lambda_4$, $\lambda_5$, and $\lambda_6$ are set to 0.01, 0.001 and 0.001.

\begin{figure*}[t]
\centering
\includegraphics[width=1.0\textwidth]{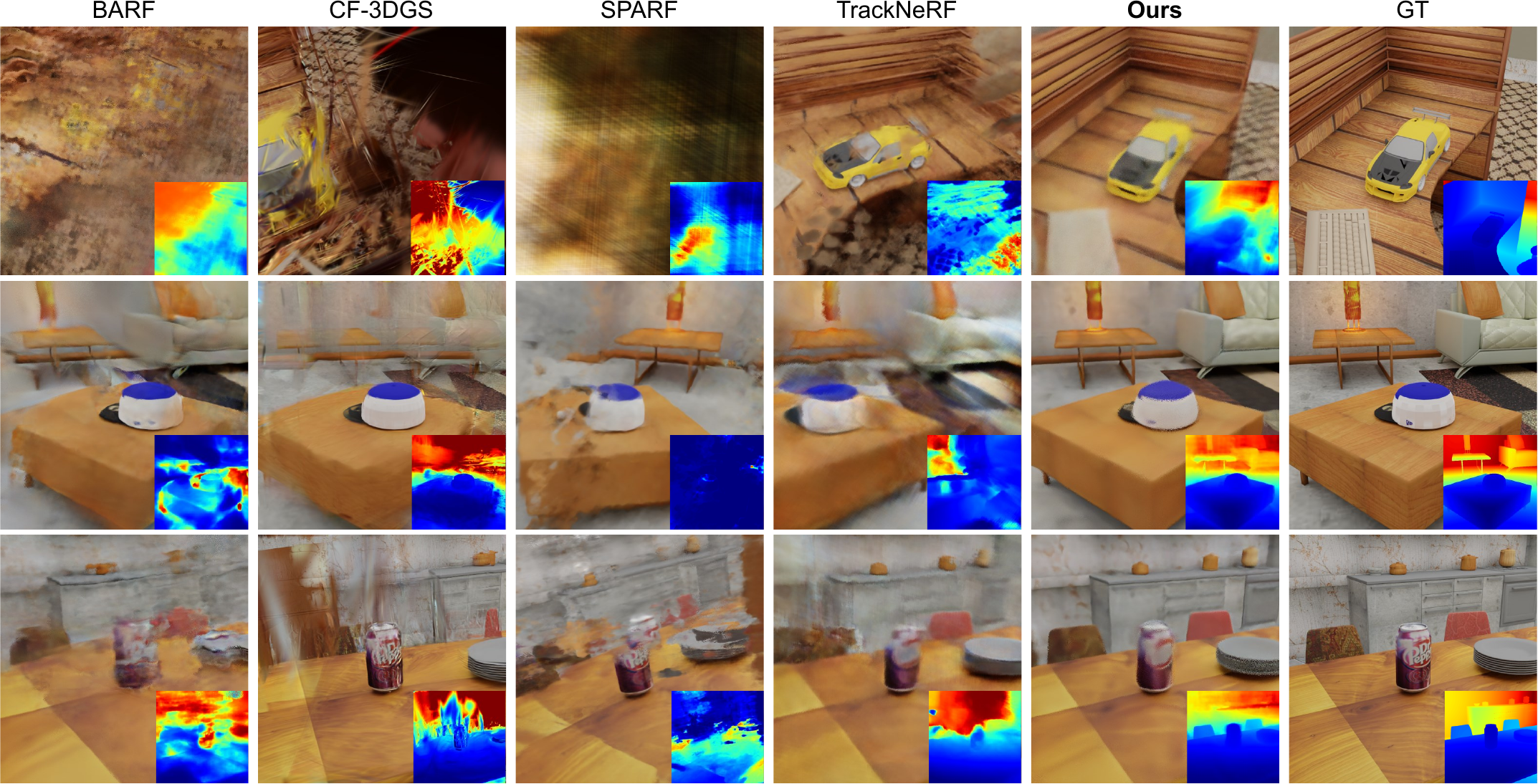}
\caption{Qualitative comparison on the ShapeScene dataset. We present novel-view RGB and depth renderings generated by baselines and our approach. The first case uses 3 input views, while the last two cases use 6 input views. Unlike the baselines, which suffer from blurriness and inaccurate scene geometry, our method generates more realistic novel views and reliable depth. We use the central objects in the scene as pose probes.} 
\label{fig:shape-img}
\end{figure*}

\begin{figure*}[tb]
    \centering
\includegraphics[width=1.0\textwidth]{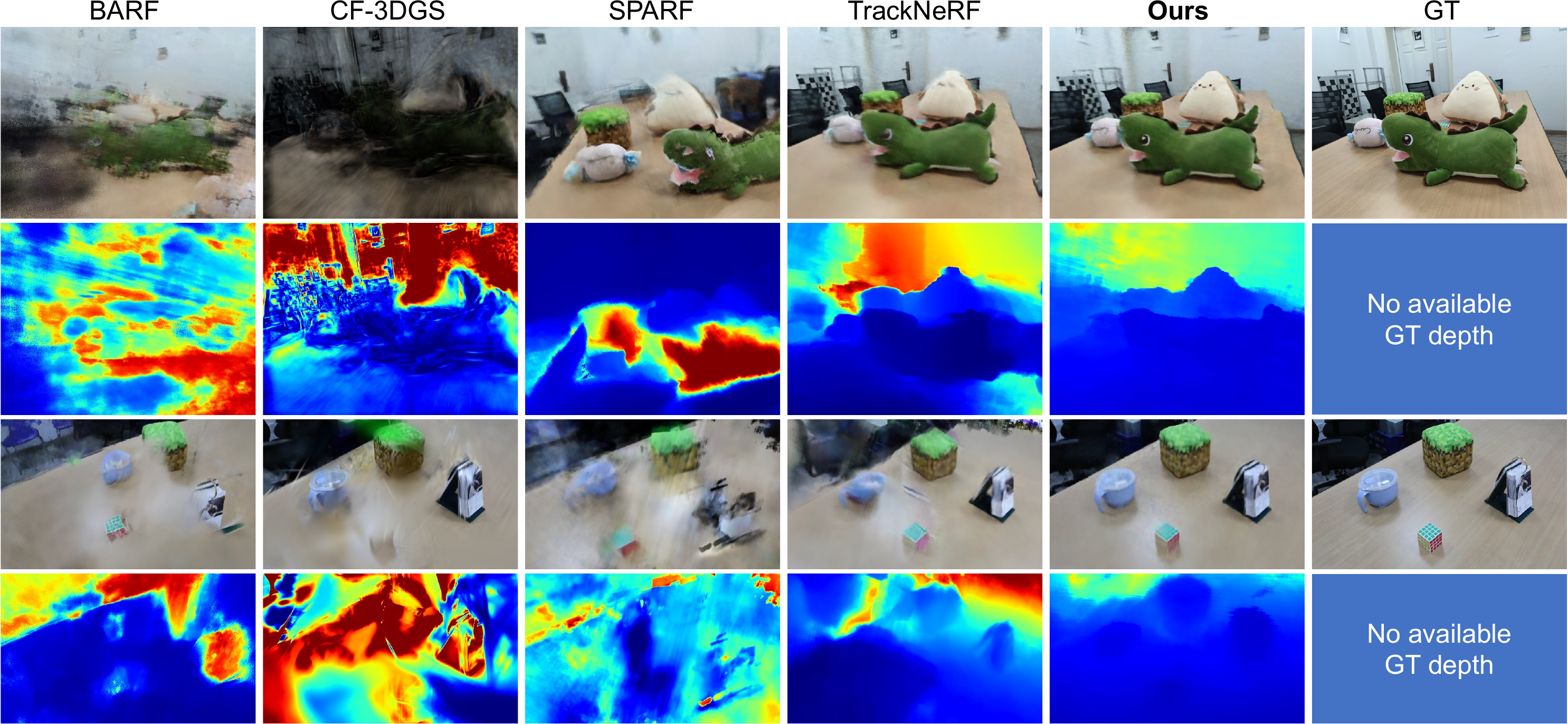
}
    \caption{
    Qualitative comparison on the ToyDesk dataset. We show novel-view renderings of the baselines and our method with 3 input views. Compared to the baselines, our method produces more detailed and realistic renderings. We use the face-shaped toy (first row) and the magic cube (second row) as pose probes.
    }
    \label{fig:toy_data}
\end{figure*}

\section{Experiments}\label{exp}
We begin with a description of the experimental settings in Sec.~\ref{sec:exp_setting}. In Sec.~\ref{sec:exp_com}, we compare our method with state-of-the-art baselines for camera pose estimation and novel view synthesis in few-view (3$\sim$6) settings across multiple benchmarks. Furthermore, in Sec.~\ref{sec:exp_abl}, we present a series of ablation studies to assess the effectiveness of key components and extensively analyze our method. Additional results are available in the supplementary videos.

\subsection{Experimental settings}\label{sec:exp_setting}
\subsubsection{Datasets} We conduct experiments on four datasets: {ShapeScene}, {ToyDesk}~\cite{yang2021objectnerf}, {DTU} ~\cite{jensen2014large}  and {Replica}~\cite{replica19arxiv}.   

We introduce ShapeScene, a texture-sparse and synthetic dataset with precise camera poses, designed as a challenging benchmark for assessing radiance fields with pose optimization. The dataset is generated using BlenderProc~\cite{Denninger2023} and consists of six scenes rendered jointly from SceneNet~\cite{handa2016scenenet} and ShapeNet~\cite{chang2015shapenet}. To demonstrate the generality of our approach, we select a diverse set of scenes and objects: two bedroom scenes, two kitchen scenes, and two office scenes from SceneNet, along with two cans, a car, a keyboard, a hat, and a sofa from ShapeNet.  Each scene includes 100 RGB images with a resolution of 512$\times$512, along with corresponding mask images,  captured around the ShapeNet object at 360$\degree$. To create a sparse-view and wide-baseline scenario, we subsample every $k^{\text{th}}$ frame and randomly select a set of consecutive training images from this subsampled set.

ToyDesk is introduced by Object-NeRF~\cite{yang2021objectnerf} and comprises two sets of posed images, each with a resolution of 480$\times$640. The scenes depict a table with various toys arranged, captured from 360\degree{} viewpoints. This setup aligns well with the design of our method, and we utilize different toys on the desk as pose probes. We subsample every $k^{\text{th}}$ frame and  a triplet of consecutive training
images.  

Replica consists of videos capturing room-scale indoor scenes. For evaluation, we select two room scenes and two office scenes. Frames are subsampled by selecting every $k^{\text{th}}$ frame, following a protocol similar to that of SPARF~\cite{truong2023sparf}, to determine training and testing images. DTU is composed of object-level scenes with wide-baseline views covering a half hemisphere. We select 8 scans that feature both foreground objects and backgrounds for our test set. The training and testing sets are separated according to the dataset splitting protocol outlined in SPARF~\cite{truong2023sparf}. In all experiments, we assume the camera intrinsic parameters are available and fixed throughout the optimization process. Our method baseline methods are evaluated on identical data splits, with default training parameters applied to the baselines.

\subsubsection{Metrics} For camera pose evaluation, we report the average rotation and translation errors as pose metrics~\cite{truong2023sparf} after aligning the optimized poses with the ground truth. For novel view synthesis, we report the PSNR, SSIM~\cite{wang2004image}, LPIPS~\cite{Zhang_2018_CVPR} (with AlexNet~\cite{krizhevsky2012imagenet}). We also present the Average metric (the geometric mean of $10^{-\mathrm{PSNR}/10}$, $\sqrt{1 - \mathrm{SSIM}}$, and LPIPS) following ~\cite{yang2023freenerf}. 

\subsubsection{Implementation details}
\label{sec-sup:impl} 
The segmentation masks of pose probes from the input images are automatically generated using Grounded-SAM~\cite{ren2024grounded,kirillov2023segany} with text prompts. In the object NeRF, the DeformNet is an MLP with three hidden layers, each containing 128 units. The SDF mapping parameters are initialized as $\beta=10$ and $\gamma=2$. Both the SDF voxel grid $V^{(\text{sdf})}$ and feature voxel grid $V^{(\text{fea})}$ have dimensions of $96^{3}$, and the step size for sampling points is 1.5 times the voxel size. The shallow MLP has four hidden layers, each with 128 units, and the feature voxel grid has 12 channels.  Coarse-to-fine positional encoding~\cite{lin2021barf} is applied, with frequency widths ranging from 0.5 to 1 for object NeRF and 0.4 to 0.7 for scene NeRF. The new frame is added every 2k iterations.  Given that too few images may not suffice to train the deformation network effectively, we fix the parameters of the deformation network while adding images. Once all images are incorporated, the entire network is optimized synchronously. We use Adam with an exponential learning rate decaying from $10^{-3}$ to $10^{-4}$ for optimizing the camera poses. The learning rates of the feature voxel grid and shallow MLP are set to $0.1$ and $0.001$. The learning of $\beta$ and $\gamma$ in the mapping function is 0.01. The object NeRF is trained for 15K iterations and the scene NeRF for 66K iterations, using a batch size of 1024 pixel rays. Camera poses are optimized only during the first 30\% of training. The entire process takes approximately 6 hours on a single NVIDIA 3090 GPU.

\begin{table*}[tb]
\centering
\caption{Quantitative evaluations on the ShapeScene. Rotation error (Rot.) is reported in degrees, while translation error (Trans.) is the ground truth scale multiplied by 100.} 
\scalebox{1.}{
\begin{tabular}{l|cc|cc|cc|cc|cc|cc}
\toprule
   &  \multicolumn{2}{c|}{Rot. $\downarrow$} & \multicolumn{2}{c}{ Trans. $\downarrow$} &  \multicolumn{2}{c|}{PSNR $\uparrow$} & \multicolumn{2}{c|}{SSIM $\uparrow$} & \multicolumn{2}{c|}{LPIPS $\downarrow$} & \multicolumn{2}{c}{Average $\downarrow$}  \\

  & 3-view & 6-view & 3-view & 6-view  & 3-view & 6-view    & 3-view & 6-view   & 3-view & 6-view   & 3-view & 6-view  \\ 
\hline

CF-3DGS~\cite{Fu_2024_CVPR} & 56.10 & 35.69 & 27.32 & 20.81 & 16.74& 18.31  &  0.49&0.65  & 0.52& 0.47&0.20  &  0.16\\ 
SCNeRF~\cite{jeong2021self} & 10.95 & 9.88 & 7.72 &  14.65& 16.39 & 16.74 & 0.51 & 0.53  & 0.58 & 0.55 & 0.21& 0.20 \\
BARF~\cite{lin2021barf} & 8.25& 13.15 & 10.53 & 10.02 & 17.95 & 18.97& 0.56 & 0.58  & 0.65 & 0.64 & 0.19 &0.17 \\ 
SPARF~\cite{truong2023sparf} &6.41 & 14.48 & 6.27& 21.45  &  18.29& 16.57   & 0.65 & 0.56   &0.45 & 0.48  & 0.16&0.19 \\

TrackNeRF~\cite{10.1007/978-3-031-73254-6_27} & 2.67& 3.42 & 5.48 & 7.11  & 18.93 &  22.57  &0.66 & 0.69   &0.41 & 0.33  & 0.15&0.10 \\

Ours  &\textbf{0.72} & \textbf{0.70 }&\textbf{1.89}&\textbf{1.06}& \textbf{23.11} &\textbf{26.08} &\textbf{0.68} & \textbf{0.79}&\textbf{0.38 }&\textbf{0.25}&\textbf{0.10}&\textbf{0.07} \\ 
\bottomrule
\end{tabular}
}

\label{tab:shapenet_tab_1} 
\end{table*}

\subsection{Comparison with State-of-the-arts}\label{sec:exp_com}

We compare our method against state-of-the-art pose-free methods, including BARF~\cite{lin2021barf}, SCNeRF~\cite{jeong2021self}, SPARF~\cite{truong2023sparf}, TrackNeRF~\cite{10.1007/978-3-031-73254-6_27}, as well as CF-3DGS~\cite{Fu_2024_CVPR}.

\begin{figure}[thb]
\centering
\includegraphics[width=0.47\textwidth]{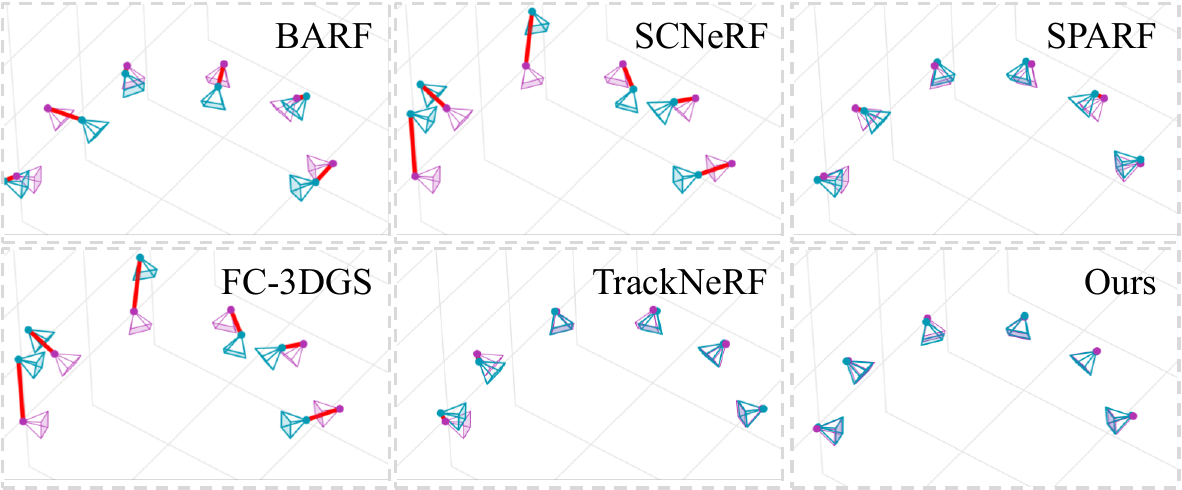} \\
\caption{Qualitative comparison for camera pose estimation.  We display the alignment of the optimized poses (in blue) with the ground truth poses (in pink), better viewed when zoomed in. }
\label{fig:pose-fig}
\end{figure}

\begin{table}[tb]
\centering

 \caption{{Quantitative comparison on the ShapeScene (3 views).} We present quantitative results optimized from GT-noisy poses. Rot. denotes errors in degrees and Trans. is the ground truth scale multiplied by 100.
 }
\scalebox{0.8}{
{
\begin{tabular}{l|cc|cccc}
\toprule
  & Rot. $\downarrow$ & Trans. $\downarrow$ & PSNR $\uparrow$ & SSIM $\uparrow$ & LPIPS $\downarrow$ & Average $\downarrow$ \\
\hline
CF-3DGS ~\cite{Fu_2024_CVPR} &  30.74 &  27.16 & 17.13 & 0.52  & 0.50 & 0.19 \\
SCNeRF~\cite{jeong2021self} &  3.59 &  6.49 & 19.65 & 0.61  & 0.43 & 0.14 \\
BARF~\cite{lin2021barf}& 10.66 & 27.43 & 16.41 & 0.52 & 0.66  & 	0.22 \\ 
SPARF~\cite{truong2023sparf} &6.04 &8.65 &  21.21  & 0.64 & 0.41 & 0.12 \\
TrackNeRF~\cite{10.1007/978-3-031-73254-6_27} &2.04 &6.27 & 21.88  & 0.65 & 0.40 & 0.11 \\
Ours & \textbf{1.31} & \textbf{1.78}  & \textbf{23.06} & \textbf{0.68} & \textbf{0.37} & \textbf{0.10} \\ 
\bottomrule
\end{tabular}
}}

 \label{tab:shapenet_tab_3}

\end{table}

\begin{figure*}[t]
\centering
\includegraphics[width=1.0\textwidth]{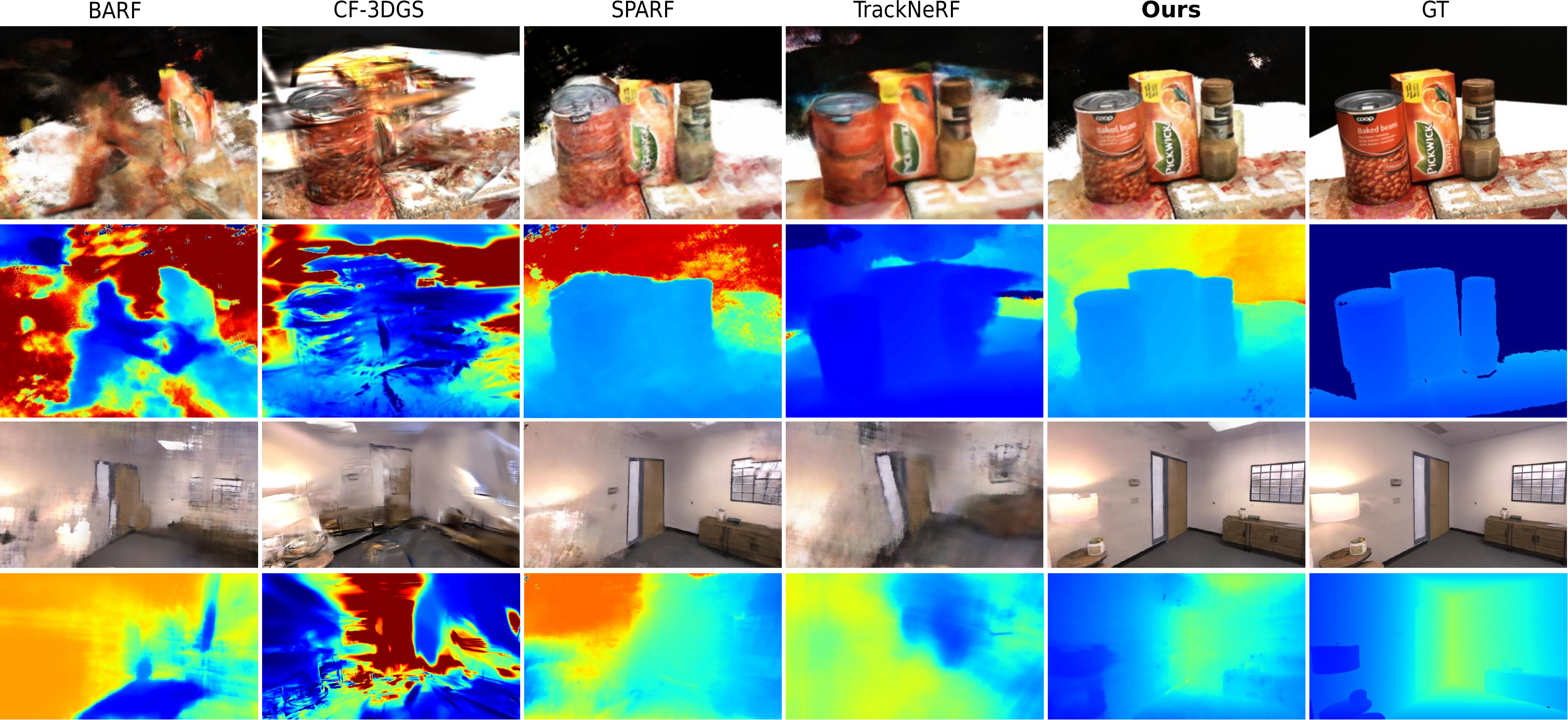}
\caption{Qualitative comparison on DTU and Replica dataset. We show novel-view RGB and depth renderings with 3 input views. Our method renders clearer details and fewer artifacts compared to other pose-free baselines. In these two examples, we use a coffee can and windows as pose probes.}
\label{fig:dtu-vis}
\end{figure*}

\subsubsection{Results on ShapeScene} 
We evaluate our method and baseline approaches using 3 and 6 input views, respectively. To ensure a fair comparison, we use camera poses derived through PnP in our method as initial poses for baselines, as these poses provide better registration than identity poses. Consequently, we also adopt this initialization method by default in subsequent experiments. The initial poses exhibit average rotation and translation errors of approximately 35$\degree$ and 70 units, respectively.  As shown in 
Tab.~\ref{tab:shapenet_tab_1} and 
Fig.~\ref{fig:shape-img}, we observe that most baseline approaches fail to accurately register poses and generate poor novel views, as they rely on good initial poses or dense input views. In Fig.~\ref{fig:pose-fig}, we present optimized poses of one scene. To further validate the robustness of our method, we experiment akin to SPARF, adding 20\% additive Gaussian noise to the ground truth poses as initial estimates. The perturbed camera poses have an average rotation and translation error of around 20$\degree$ and 50 units, respectively. Quantitative results are shown in Tab.~\ref{tab:shapenet_tab_3}. Both BARF and CF-3DGS face persistent challenges in optimizing camera poses, leading to subpar rendering quality. The geometric projection losses employed by SCNeRF, SPARF, and TrackNeRF contribute to the improvement of camera pose estimation. However, they still face challenges in accurately recovering the precise poses. In contrast, our method achieves more accurate pose estimation and more realistic renderings, both from scratch and noisy poses, resulting in higher-quality novel views.

\subsubsection{Results on ToyDesk} 

Table~\ref{tab:toy_desk_3} presents the quantitative results on the ToyDesk dataset using three input images. To ensure consistency in the evaluation, the initial camera poses for all baselines are derived from PnP using pose probes. These initial poses have average rotation and translation errors of approximately 11.8$\degree$ and 9.3, respectively. Notably, CF-3DGS, SCNeRF, and BARF fail to register the camera poses effectively. TrackNeRF performs better than SPARF, recovering more accurate camera poses from highly noisy initial estimates. The visual results are illustrated in Fig.~\ref{fig:toy_data}, where our method achieves lower camera pose errors and improved rendering quality. These improvements can be attributed to two critical factors: the strong geometric constraints provided by pose probes in the object branch and the global multilayer feature-metric consistency enforced by the scene branch.

\begin{table}[tb]
\centering

 \caption{{Quantitative comparison on the ToyDesk (3 views).} Rot. denotes errors in degrees and Trans. is the ground truth scale multiplied by 100.
 }

\scalebox{0.8}{
{
\begin{tabular}{l|cc|cccc}
\toprule
  & Rot. $\downarrow$ & Trans. $\downarrow$ & PSNR $\uparrow$ & SSIM $\uparrow$ & LPIPS $\downarrow$ & Average $\downarrow$ \\
\hline
CF-3DGS ~\cite{Fu_2024_CVPR} &  53.59 &  21.66& 12.89 & 0.59  & 0.73 & 0.29 \\
SCNeRF~\cite{jeong2021self} &  14.64 &  17.86 & 13.88 & 0.59  & 0.71 & 0.26 \\
BARF~\cite{lin2021barf}& 20.38 &  19.12  & 13.17 & 0.57  & 0.76&	0.29 \\ 
SPARF~\cite{truong2023sparf} &7.71 & 8.69 &   15.29 & 0.62 & 0.66 & 0.23 \\
TrackNeRF~\cite{10.1007/978-3-031-73254-6_27} &0.40 &0.65 & 18.46  & 0.69 & 0.49 & 0.15 \\
Ours & \textbf{0.29} & \textbf{0.24}  & \textbf{19.40} & \textbf{0.71} & \textbf{0.43} & \textbf{0.14} \\ 
\bottomrule
\end{tabular}
}}

 \label{tab:toy_desk_3}

\end{table}

\begin{table}
\centering
\caption{Quantitative comparison on the DTU (3 views). Rot. is in degrees and Trans. is the ground truth scale multiplied by 100.
}
\scalebox{0.8}
{
\begin{tabular}{l|cc|cccc}
\toprule
  & Rot. $\downarrow$ & Trans. $\downarrow$ & PSNR $\uparrow$ & SSIM $\uparrow$ & LPIPS $\downarrow$ & Average $\downarrow$ \\
\hline
CF-3DGS~\cite{Fu_2024_CVPR} &20.61&22.45& 10.84 & 0.39 & 0.53
&0.32   	 \\
SCNeRF~\cite{jeong2021self} & 11.63&19.13 &12.62& 0.45 &0.57	& 0.28 \\ 
BARF~\cite{lin2021barf} &	15.37&45.13 &10.08& 0.39 &0.60	& 0.35	 \\ 

SPARF~\cite{truong2023sparf} &3.46 & 6.48 & 16.79
& 0.64  & 0.30& 0.16  	 \\
TrackNeRF~\cite{10.1007/978-3-031-73254-6_27}  &1.83 & 4.07& 17.58 &
0.68 & \textbf{0.26}  & 0.14	 \\

Ours & \textbf{1.27} & \textbf{3.82} & \textbf{17.92}& \textbf{0.69} & \textbf{0.26} & \textbf{0.13} \\
\bottomrule
\end{tabular}
}

\label{tab:result_dtu} 
\end{table}

\subsubsection{Results on DTU and replica}  We test the DTU and replica datasets with 3 input views and compare our method with state-of-the-art techniques.  
The PnP camera poses are used as the initial poses for baselines to ensure a fair comparison. The visual results are presented in Fig.~\ref{fig:dtu-vis}, the quantitative results are presented in Tab.~\ref{tab:result_dtu} and Tab.~\ref{tab:replica_3}, our method performs better in pose estimation and novel-view synthesis. All baselines suffer from blurriness and inaccurate scene geometry as they rely on strong pose initialization assumptions, while our approach produces closer results to GT thanks to the pose probe constraint.

\begin{table}[th]
\centering
 \caption{{Quantitative comparison on the Replica (3 views).} Rot. denotes errors in degrees and Trans. is the ground truth scale multiplied by 100.
 }
\scalebox{0.8}{
{
\begin{tabular}{l|cc|cccc}
\toprule
  & Rot. $\downarrow$ & Trans. $\downarrow$ & PSNR $\uparrow$ & SSIM $\uparrow$ & LPIPS $\downarrow$ & Average $\downarrow$ \\
\hline
CF-3DGS ~\cite{Fu_2024_CVPR} &13.09   & 19.7  &  14.84&  0.65 &  0.49 & 0.21 \\
SCNeRF~\cite{jeong2021self} & 3.91  & 8.25  & 18.58 & 0.68  & 0.37 & 0.14\\
BARF~\cite{lin2021barf}& 5.94 &22.74 & 15.17 & 0.66 & 0.42  & 0.20\\ 
SPARF~\cite{truong2023sparf} &3.25 & 6.39 &  22.18  & 0.76 & 0.30 & 0.10 \\
TrackNeRF~\cite{10.1007/978-3-031-73254-6_27} &2.27 & 3.19 &  23.79 & 0.81 & 0.25 & 0.07 \\
Ours & \textbf{0.38} & \textbf{1.05}  & \textbf{25.05} & \textbf{0.85} & \textbf{0.17} & \textbf{0.06} \\ 
\bottomrule
\end{tabular}
}}

 \label{tab:replica_3}

\end{table}

\begin{table*}
\centering
\caption{ 
Effect of key modules. We evaluate the impact of different modules by individually disabling them on the ToyDesk dataset.
Rot. is in degrees and Trans. is the ground truth scale multiplied by 100.
}
\scalebox{1.}
{
\begin{tabular}{c|cc|cccc | c}
\toprule
 & {Rot. $\downarrow$} & {Trans. $\downarrow$} & {PSNR $\uparrow$} & {SSIM $\uparrow$} & {LPIPS $\downarrow$} & {Average $\downarrow$}   & {Times (hrs)}
\\
\hline

w/o PnPInit &19.03 & {23.71} & {14.06}  &  {0.57} & {0.74} & {0.26} & {6.1}\\

w/o Incre. & 4.62 & 6.03 & 16.39  &  0.66 & 0.60 &0.20 & 6.1\\

w/o $\mathcal{L}_\text{Fea}$& 0.43 &  0.83&  18.41  & 0.69 & 0.45 & 0.15 & 5.1\\

 w/o $\mathcal{L}_{\text{Dist}}$ & 0.32 & 0.36 & 19.28 & 0.71 & 0.44 & 0.14 & 5.8\\ 

 w/o $\mathcal{L}_\text{GeoDist}$& 0.56 & 0.62 & 19.02 & 0.70 & 0.44 & 0.14 & 6.1\\

w/o $\mathcal{L}_\text{GeoSce}$ & 2.53 & 5.92 & 17.14 & 0.61 & 0.51 & 0.18 & 5.6 \\ 

w/o $\mathcal{L}_\text{GeoBoth}$ & 17.43 & 15.73 & 14.58 & 0.59 & 0.73 & 0.25 & 5.3\\ 

w/o DeformNet & 2.14 & 3.56 & 17.48  & 0.62 & 0.51 &0.17 & 5.7\\ 
Full Model & \textbf{0.29} & \textbf{0.24} & \textbf{19.40} & \textbf{0.71} & \textbf{0.43} & \textbf{0.14} & 6.1\\

\bottomrule
\end{tabular} 
}

\label{tab:abl}
\end{table*}

\begin{figure}[tbh]
\centering
\includegraphics[width=0.48\textwidth]{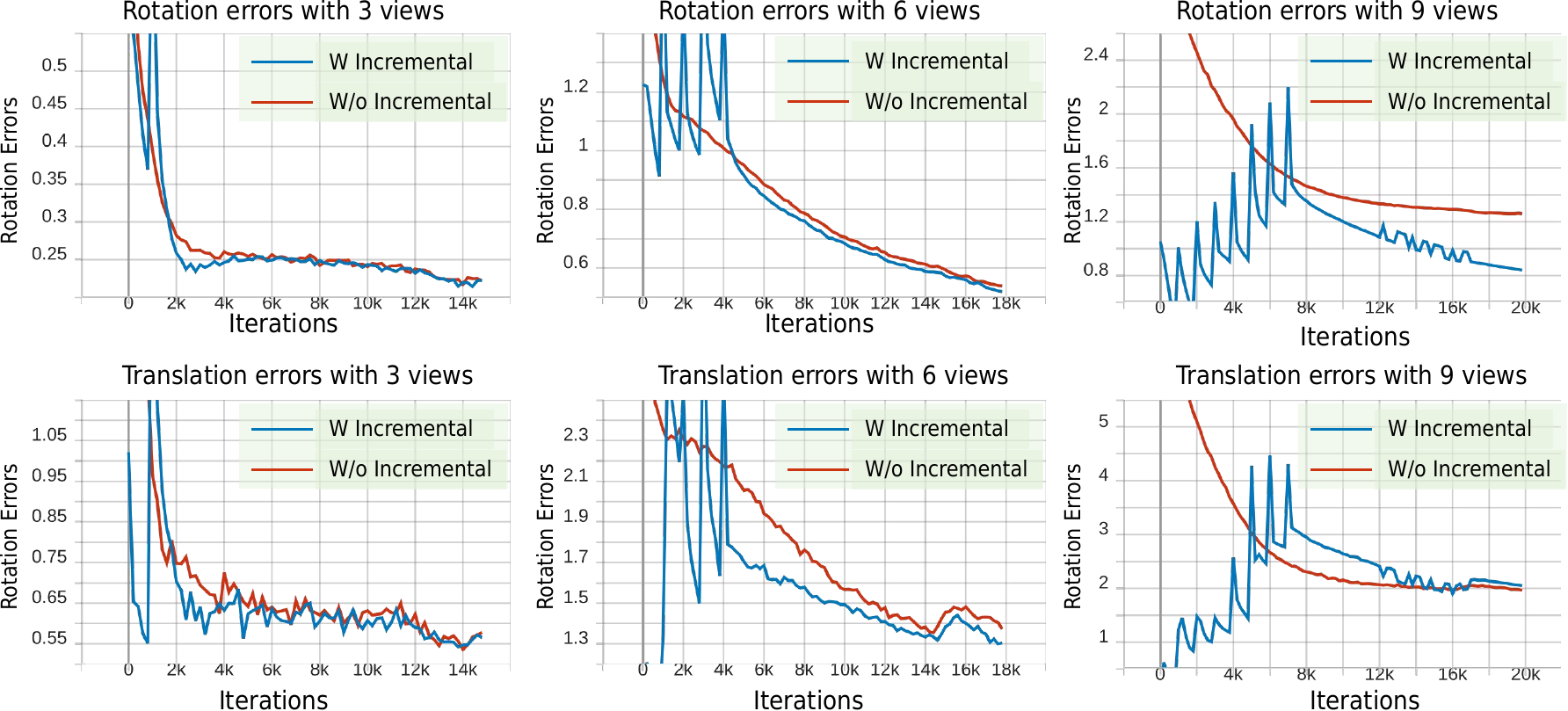} \\
  \caption{Rotation and translation errors with or without incremental pose optimization. The blue curve (\textbf{Ours}) represents the inclusion of incremental pose optimization, and the red curve indicates that incremental pose optimization is disabled. It is observed that incremental pose optimization enables camera poses to converge sequentially and eventually to a better solution.}
    \label{fig:effct_pnp}
\label{fig:deform_mesh}
\end{figure}

\subsection{Ablations and analysis}  \label{sec:exp_abl}
In this section, we conduct ablation studies and analyze the robustness of our approach. The ToyDesk dataset, featuring daily scenes and multiple objects as pose probes, is chosen as the primary dataset for our ablation experiments.
\subsubsection{Effectiveness of proposed components}   \label{sec:comp_abl}
As shown in Tab.~\ref{tab:abl}, we evaluate the effectiveness of key modules by conducting ablation studies using three input views on the ToyDesk dataset.
\begin{itemize}

\item We establish 2D-3D correspondences and compute the initial camera poses using PnP with RANSAC. This procedure yields a high-quality initial solution, significantly reducing the risk of converging to local minima. As demonstrated in the first row of Tab~\ref{tab:abl}, using identical camera poses as the initial guess leads to a substantial performance drop.

\item As more images are added, errors in the camera pose estimates accumulate, leading to poor initial poses and potentially trapping the optimization in local minima. To mitigate this issue, we propose an incremental pose optimization strategy that calculates the initial poses of new frames based on the previously optimized poses. Omitting this strategy results in a substantial drop in model performance. To provide an intuitive demonstration of this strategy, Fig.~\ref{fig:effct_pnp} illustrates the pose error curves, comparing results with and without incremental optimization. Note that our method shows higher initial errors during the progressive integration of new camera views. When a new frame is added to the optimization loop, its initial pose estimate may have relatively large errors due to challenging viewpoints. However, as seen in the later stages of the error curves, the incremental strategy allows the camera poses to converge sequentially, ultimately achieving better final accuracy.

\item In the scene branch, we employ multi-layer feature consistency (Eqn.~\ref{eq:fea-con}), and density weighting to achieve a compact distribution (Eqn.~\ref{eq:dist_loss}). The former enhances the registration of camera poses, while the latter facilitates producing higher-quality images with minimal impact on pose optimization.

\item  The geometric consistency loss (Eqn.~\ref{eq:geo-cons}) plays a crucial role in guiding camera pose optimization.  As shown in Tab.~\ref{tab:abl}, the geometric consistency terms in the object branch ($\mathcal{L}_\text{GeoObj}$) and in the scene branch ($\mathcal{L}_\text{GeoScn}$) both contribute to learning better camera poses. Combining two terms ($\mathcal{L}_\text{GeoBoth}$) leads to the highest camera pose accuracy. Omitting these constraints results in a noticeable decline in performance, as inaccurate poses degrade the quality of novel view synthesis.

\item DeformNet plays a crucial role in our framework by refining the shape of probe objects and enhancing geometric constraints. This refinement enables more accurate camera pose estimation, resulting in higher-quality novel view synthesis.

\item The final column of Tab.~\ref{tab:abl} presents the training time for our method and its ablation studies. Our approach requires approximately 6 hours per scene. In comparison, SPARF and TackNeRF typically take around 10 hours, while BARF and SCNeRF require about 4 hours. CF-3DGS demonstrates the highest efficiency, needing only a few minutes.

\end{itemize}

\begin{figure}[tbh]
\centering
\includegraphics[width=0.48\textwidth,height=0.2\textheight, keepaspectratio=false]{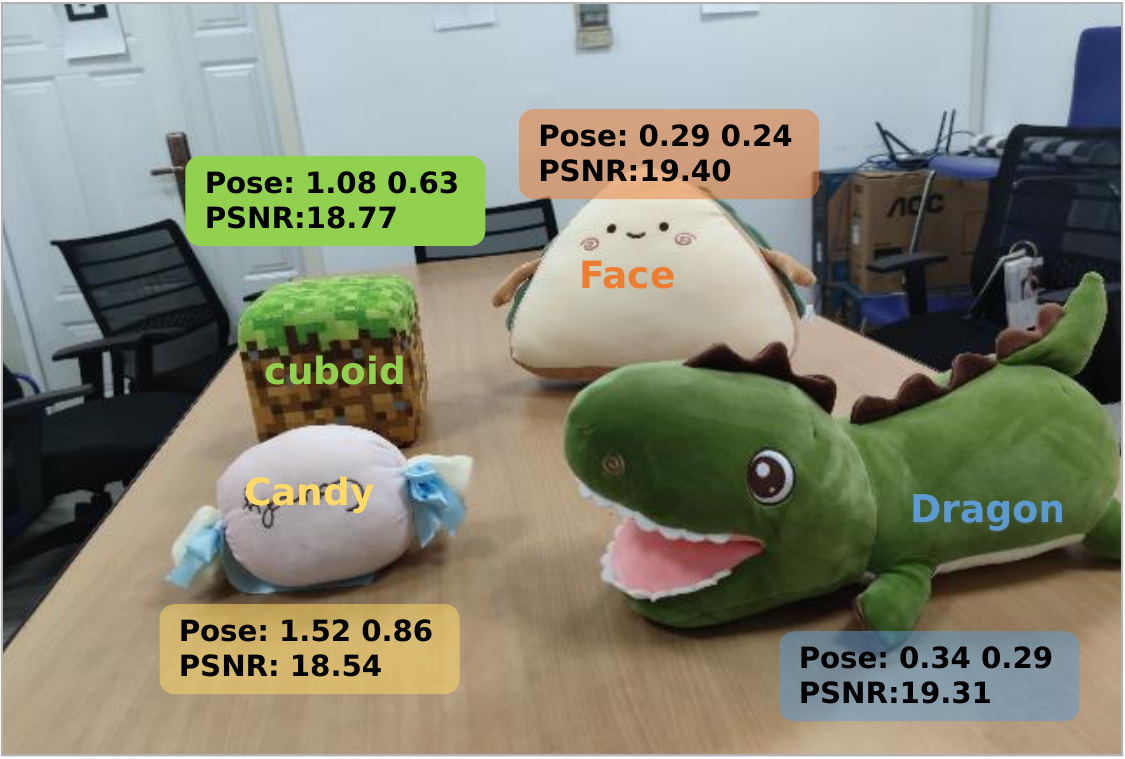} \\
  \caption{Evaluation on one scene of ToyDesk with different pose probes. We use the Candy, Face, and Dragon toys in the scene as pose probes, respectively.}
\label{fig:eff_probe}
\end{figure}

\subsubsection{Impact of different pose probes} To investigate the insensitivity to different pose probes, we utilize a scene from ToyDesk containing multiple partially observed objects. Four toys in the scene are alternately used as pose probes, with all shapes initialized as cuboids. The optimized pose errors and average PSNR values for novel view synthesis are reported in Fig.~\ref{fig:eff_probe}, showing that all pose probes perform effectively.

\begin{table*}[tbh]
\centering
\caption{Performance of COLMAP-based methods and ours. SR is the success rate of recovering all camera poses, while matches are the number of matching pairs. ``Ours-50\%" and ``Ours-20\%" indicate that 50\%  and 20\% of the matches are used.}
{
\begin{tabular}{l|cccc|cccc}
\toprule
 \multirow{2}{*}{Pose init.}  & \multicolumn{4}{c|}{Sparse views} & \multicolumn{4}{c}{Dense views} \\

 & Rot. $\downarrow$ & Trans. $\downarrow$ & SR $\uparrow$  & Matches & Rot. $\downarrow$ & Trans. $\downarrow$ & SR $\uparrow$ & Matches \\
\hline

 COLMAP~\cite{7780814} & - & - &  0.0\% & 202 & 3.38  & 8.82 & 83\% & 2271 \\ 
  COLAMP-SP-SG~\cite{detone2018superpoint,sarlin2020superglue} & 10.24 & 11.61 & 33\% & 499 & 13.58 & 2.32 & 100\%& 3208 \\

  \hline

 { Ours-20\%} &  {4.27} &  {7.83}  &  {100\%} &  {31} &  {3.92} &  {5.03} & 
  {100\%}  &
   {157}\\
  
  Ours-50\% & 1.97 & 2.72  & 100\% & 137 & 1.48 & 2.91 & 100\%  &392\\

Ours & \textbf{0.72} & \textbf{1.89}  &  100\% & 274 & \textbf{0.70} & \textbf{1.06} & 100\%  & 783 \\

\bottomrule
\end{tabular}
}

\label{tab:ours_colmap} 
\end{table*}

\begin{table}[tb]
\centering
\caption{Performance of different pose initialization strategies on ShapeScene dataset (3 views).}
\resizebox{0.48\textwidth}{!}{
\begin{tabular}{l|cc|cccc}
\toprule
  Pose init.&  Rot. $\downarrow$ & Trans. $\downarrow$ & PSNR $\uparrow$ & SSIM $\uparrow$ & LPIPS $\downarrow$ & Average  $\downarrow$  \\
\hline

Identical  & 1.11 & 3.15& 22.82 &0.67  & 	0.38 &  0.104 \\ 
30\% noise  &2.84  & 7.16 &22.25 & 0.63 &0.46& 0.119 \\ 
25\% noise & 0.81& 1.82  & 22.91 & 0.67 & 0.40 & 0.105 \\ 
15\% noise & 0.80& 2.30  & 22.59 & 0.66 & 0.39 & 0.108 \\ 
PnP  &\textbf{0.72} & \textbf{1.89}  & \textbf{23.11} & \textbf{0.68} & \textbf{0.38} & \textbf{0.102} \\ 
\bottomrule
\end{tabular}
}

\label{tab:pose_init}
\end{table}

\subsubsection{Robustness matching pairs and  on initial poses} 

We compare the robustness of COLMAP~\cite{7780814} and our PnP method by categorizing the data into sparse (3 views) and dense (6 views) splits, using ShapeScene as the dataset for this evaluation. As shown in Tab.~\ref{tab:ours_colmap}, the state-of-the-art COLMAP with SuperPoint and SuperGlue (COLMAP-SP-SG) frequently fails in the sparse view split due to insufficient feature pairs for pose initialization. Unlike the SfM pipeline, PnP eliminates the need for the reconstruction process to recover camera poses, relying on fewer features. 
Our results demonstrate that the proposed method remains effective even when using only half of the matching pairs, highlighting its reduced dependency on pose matching compared to COLMAP. PnP operates reliably with significantly fewer feature pairs, making it suitable for both sparse and dense view settings.  It is worth mentioning that the COLMAP poses are further refined using SPARF~\cite{truong2023sparf}. Although our method employs PnP to compute the initial poses of new frames, it is not strictly necessary. PnP initialization primarily accelerates pose convergence by providing a more accurate starting point. To validate this, we conducted experiments with 3 input views using various pose initialization strategies, as shown in Tab.~\ref{tab:pose_init}. Our method achieves comparable performance when using the previous frame's pose as initialization (identical poses) and remains robust even with substantially noisy poses.

\begin{figure}[tbh]
\centering
\includegraphics[width=0.5\textwidth]{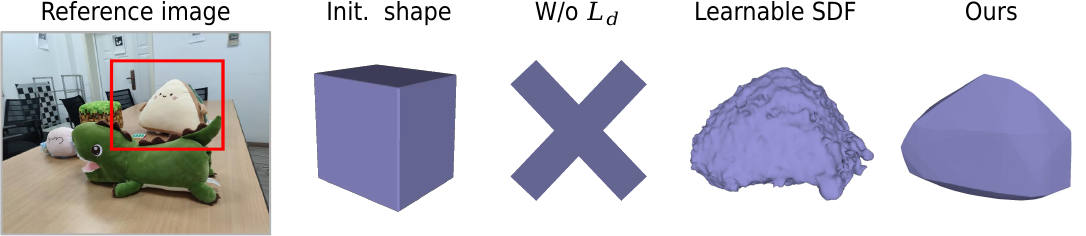} \\
\caption{ 
The impact of different strategies on the object shape is demonstrated. The third column illustrates the shape of removing the deformation regularization loss, while the fourth column shows the result of directly optimizing the SDF grid instead of utilizing a DeformNet.
}
\label{fig:deform_abl}

\end{figure}

\begin{figure}[tbh]
\centering
\includegraphics[width=0.5\textwidth]{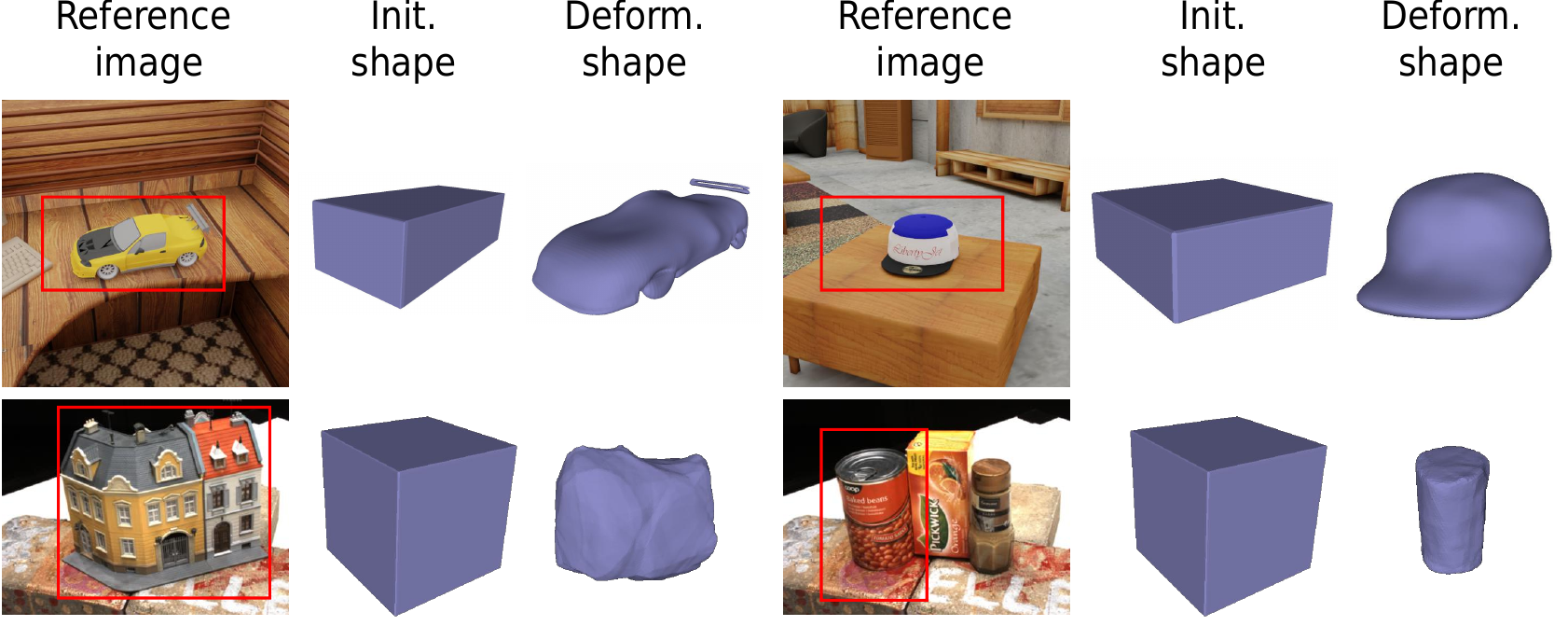} \\
\caption{ We present the initial and deformed shapes of the pose probe using the ShapeScene and DTU datasets. The first row corresponds to six input views, while the second row corresponds to three input views. The pose probe within the scene is highlighted by a red box.
}
\label{fig:deform_mesh}

\end{figure}

\begin{table}

\centering
\caption{The effect on Performance with noisy masks on ToyDesk dataset with 3 input views.}
\resizebox{0.48\textwidth}{!}{
\begin{tabular}{l|cc|cccc}
\toprule
  Method & Rot. $\downarrow$ & Trans. $\downarrow$ & PSNR $\uparrow$ & SSIM $\uparrow$ & LPIPS $\downarrow$ & Average $\downarrow$ \\
\hline
mask Erosion & 0.82 & 0.73 & 18.51 & 0.70 & 0.44 & 0.15
\\
mask Dilation & 0.35 & 0.41 & 19.15 & 0.71 &0.44 & 0.15
\\
SAM mask & \textbf{0.29} & \textbf{0.24}  & \textbf{19.40} & \textbf{0.71} & \textbf{0.43} & 0.14 \\ 
\bottomrule
\end{tabular}
}

\label{tab:noise_mask} 
\end{table}

\subsubsection{Robustness when mask noises exist}
In our pipeline, Grounded-SAM~\cite{ren2024grounded,kirillov2023segany} is employed with text prompts to generate masks for the object branch, which inevitably contain noise.  To evaluate robustness, we performed ablation studies by introducing noise to the masks, as well as dilating and eroding them by 16\%, as detailed in Tab.~\ref{tab:noise_mask}. The results demonstrate that our method exhibits good resilience to mask noise, maintaining comparable performance despite these perturbations.

\subsubsection{Object reconstruction}  \label{sec:obj_rec}

\begin{figure}[tbh]
\centering
\includegraphics[width=0.48\textwidth]{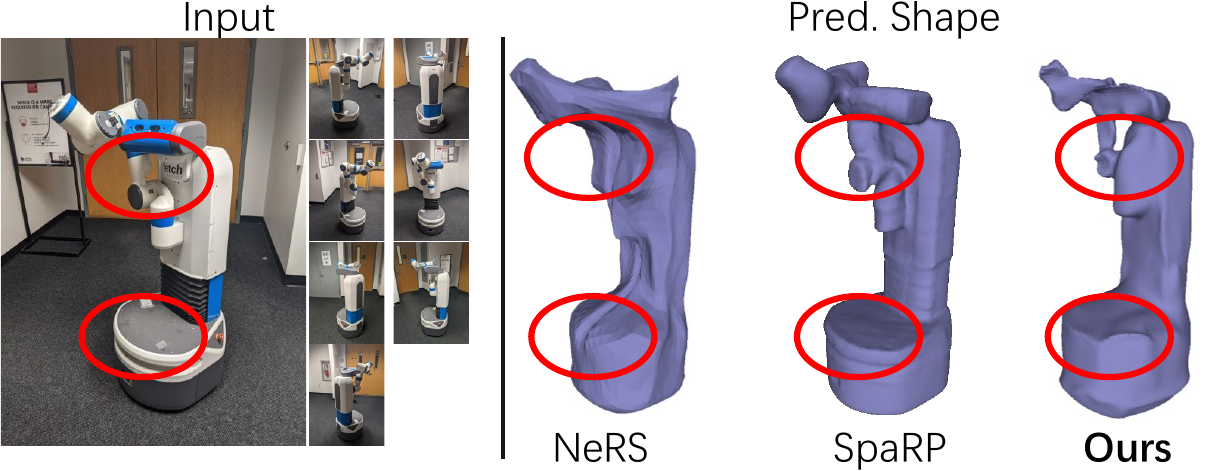} \\
\caption{
Comparison of our method with NeRS and spaRP in geometric reconstruction.
}
\label{fig:ners}
\end{figure}

Although our method does not focus on the reconstruction quality of the pose probes, we showcase the geometry of the objects after the deformation network to better understand our approach in Fig.~\ref{fig:deform_abl}. The deformation regularization (Eq.\ref{eq:deform_reg}) serves as a crucial constraint for DeformNet. Without this regularization, the shape of the probe object collapses, as the SDF field becomes entirely negative, leading to an empty shape. Another approach to refine the object's shape is directly learning the SDF voxel grid, similar to Voxurf~\cite{wu2022voxurf}. However, this method often results in significant artifacts in the reconstructed shape under sparse-view settings. In contrast, the implicit DeformNet offers a more stable and smoother shape. As shown in Fig.~\ref{fig:deform_mesh}, it can be observed that even with few input views and a limited number of optimization steps, our object branch achieves a reasonable geometry. We compare our reconstruction approach with NeRS~\cite{zhang2021ners}, as both methods follow a similar strategy of learning a deformation field from an initial shape.  We compare our reconstruction approach with NeRS~\cite{zhang2021ners}, as both methods adopt a similar strategy of learning a deformation field from an initial shape. As shown in Fig.~\ref{fig:ners}, using the dataset provided by NeRS, our method effectively addresses non-convex objects, yielding more accurate reconstructions. Additionally, we include the result of SpaRP~\cite{xu2025sparp} as a reference, which represents the current state-of-the-art method for reconstructing objects from sparse and unposed views.

\subsubsection{Performance when increasing views}
\begin{figure}[th]
\centering
\includegraphics[width=0.48\textwidth]{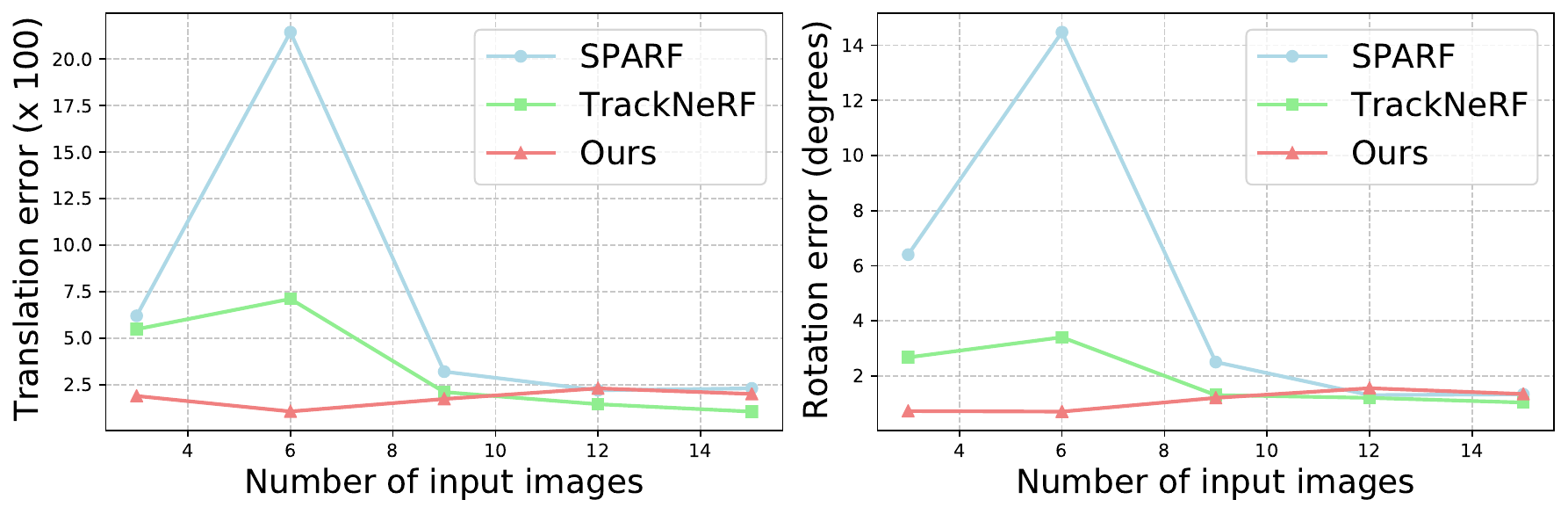} \\
  \caption{Translation and rotation errors under different input views on the ShapeScene dataset.}
\label{fig:pose_error_curve}
\end{figure}

As illustrated in Fig.~\ref{fig:pose_error_curve}, our method demonstrates a clear advantage when the number of camera views is limited ($\le$6). As the number of views increases further, our method achieves comparable performance to other approaches.

\begin{figure}[tbh]
\centering
\includegraphics[width=0.48\textwidth]{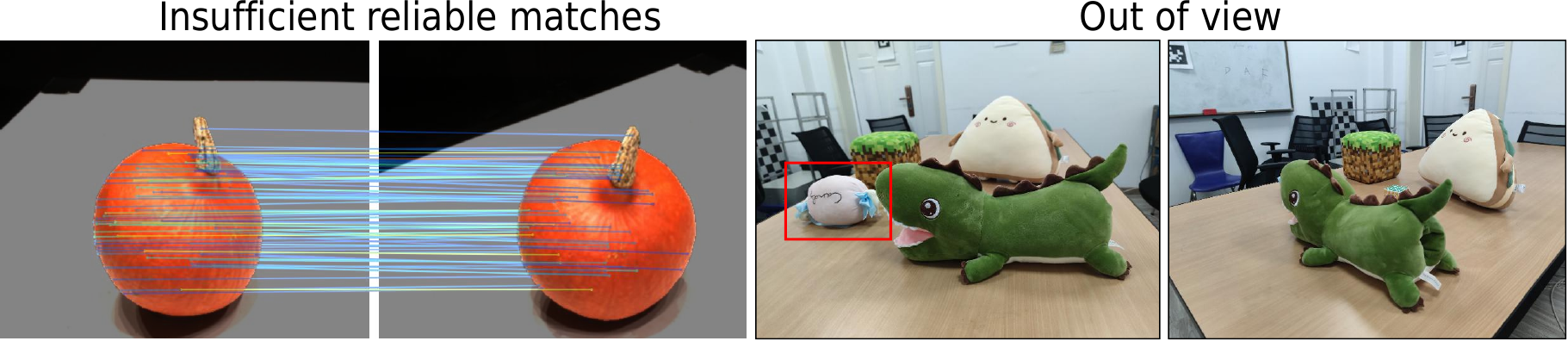} \\
  \caption{The failure of our method occurs when the pose probe either has too few reliable matches or is no longer visible in the subsequent image sequences.}
\label{fig:fail_case}
\end{figure}

\subsection{Failure cases}
Our approach depends on extracting correspondences from images of the pose probe to initialize camera poses and enforce consistency constraints during subsequent optimization. This process may fail in two common scenarios, as illustrated in Fig.~\ref{fig:fail_case}. First, if the pose probe image contains a limited number of features or exhibits symmetry, feature matching algorithms may fail to establish a sufficient number of reliable correspondences. Second, failure can occur if the pose probe is not visible within the camera's field of view, due to occlusion or substantial motion.


\section{Conclusion}\label{conclusion}
We propose PoseProbe, a novel pipeline that leverages generic objects as camera pose probes for joint pose-NeRF training, tailored for challenging scenarios of few-view and large-baseline, where COLMAP is infeasible.  By utilizing a hybrid SDF representation of the pose probe, we incrementally acquire initial camera poses, which are subsequently refined through joint optimization with object NeRF and scene NeRF.  Multi-view geometry and multi-layer feature consistency are explored to enable the accurate joint optimization of NeRF and camera poses. Our method achieves state-of-the-art performance in pose estimation and novel view synthesis across multiple challenging datasets.  

\textbf{Limitations and future work.} 
 Although the pose probe provides strong constraints for pose optimization and significantly improves accuracy, the reliance of our approach on 2D correspondences may face challenges in scenarios involving featureless objects or SAM segmentation failures. Furthermore, the pose probe necessitates persistent visibility across all camera viewpoints, which imposes practical constraints in complex environments. These limitations motivate future exploration of automated multi-calibration object utilization within scenes to enhance robustness. Another promising direction involves joint optimization of camera intrinsics and extrinsics for scenarios with unknown or noisy intrinsic parameters.

\bibliographystyle{IEEETran}
\bibliography{ref}

\vspace{-11pt}
\begin{IEEEbiography}[{\includegraphics[width=1in,height=1.20in,clip,keepaspectratio]{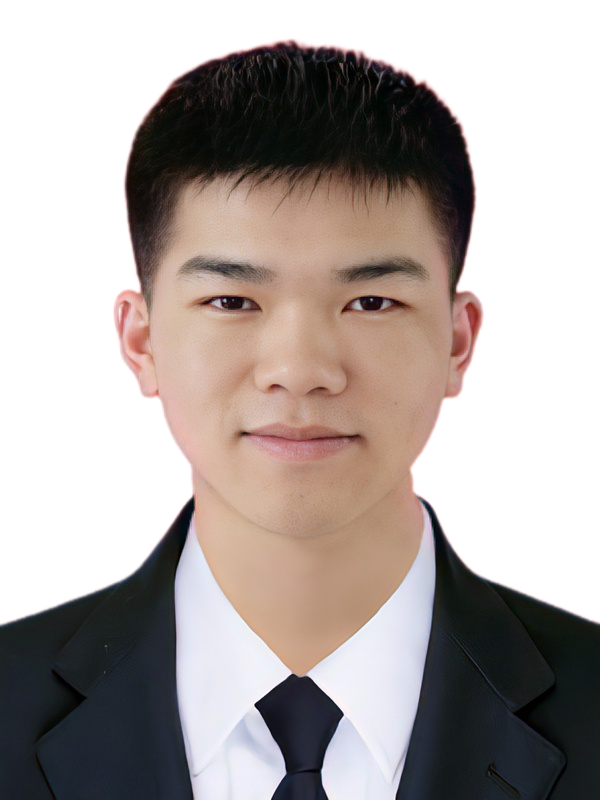}}]{Zhirui Gao}
received his B.E. degree in computer science and technology from China University of Geosciences, Wuhan in 2021. He
is now a Ph.D. Student at the National University of Defense Technology, China. His research interests include computer graphics and 3D vision.
\end{IEEEbiography}

\vspace{-11pt}
\begin{IEEEbiography}[{\includegraphics[width=1in,height=1.20in,clip,keepaspectratio]{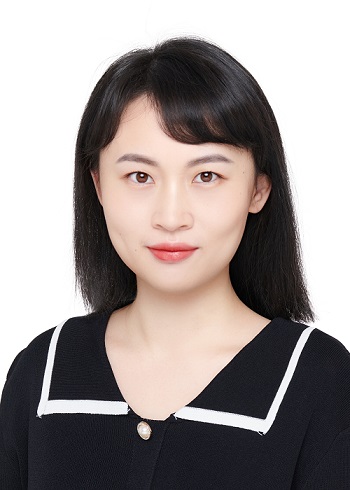}}]{Renjiao Yi} is an Associate Professor at the School of Computer Science, National University of Defense Technology. She received a Bachelor’s degree in Computer Science from National University of Defense Technology, China, in 2013, and a Ph.D. from Simon Fraser University, Burnaby, BC, Canada, in 2019. Her research interests include inverse rendering, image relighting, scene reconstruction in 3D vision and graphics. 
\end{IEEEbiography}

\vspace{-11pt}
\begin{IEEEbiography}[{\includegraphics[width=1in,height=1.20in,clip,keepaspectratio]{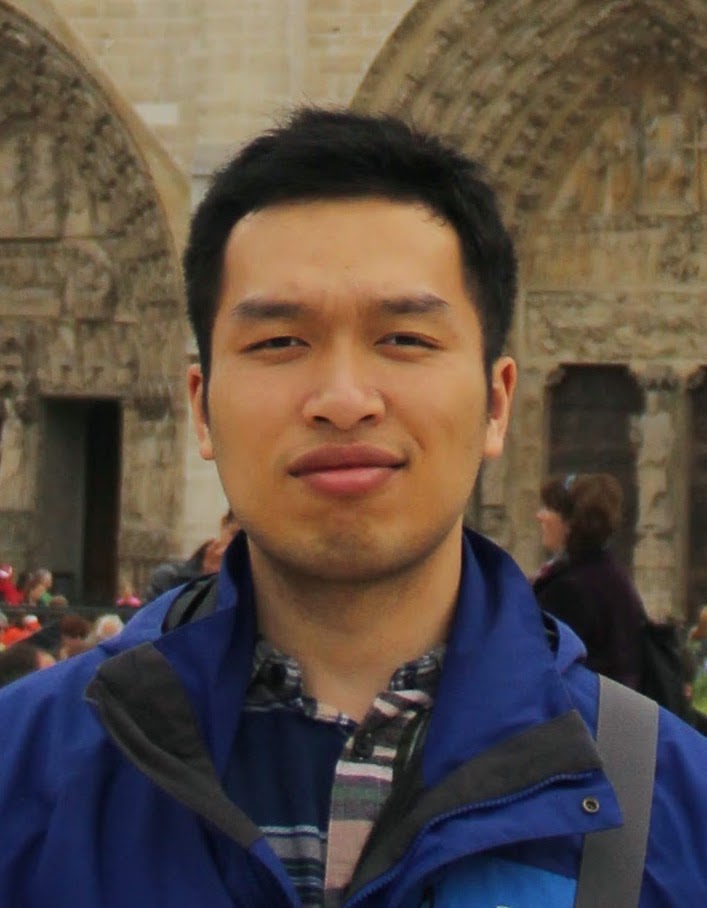}}]{Chenyang Zhu} is an Associate Professor at the School of Computer Science, National University of Defense Technology (NUDT).  He holds both a Bachelor’s and a Master’s degree in Computer Science from NUDT, earned in June 2011 and December 2013, and completed his PhD at Simon Fraser University. His research focuses on computer graphics, 3D vision, robotic perception, and navigation.
\end{IEEEbiography}

\vspace{-11pt}
\begin{IEEEbiography}[{\includegraphics[width=1in,height=1.20in,clip,keepaspectratio]{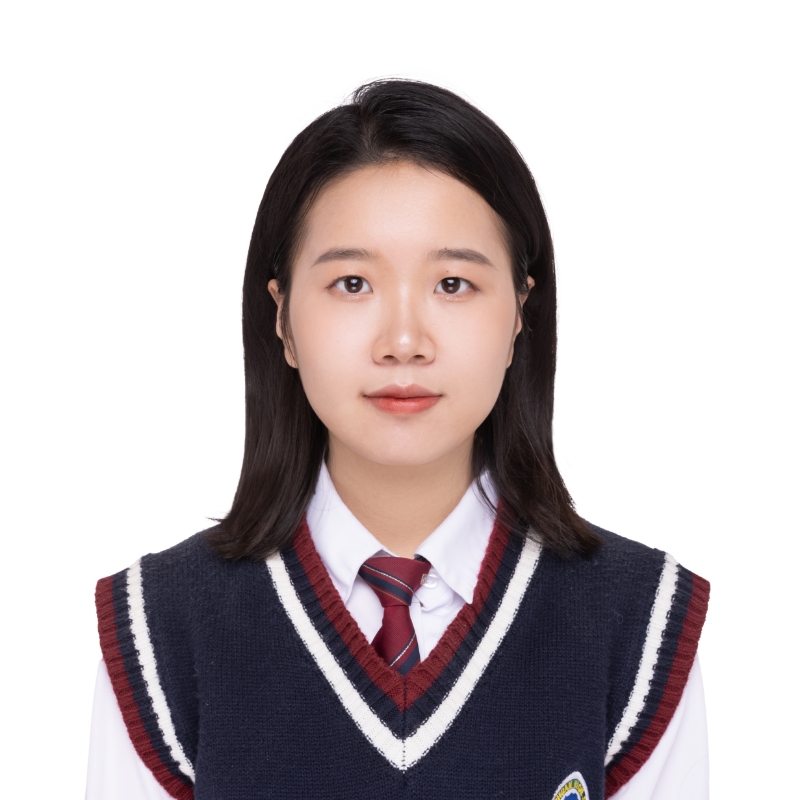}}]{Ke Zhuang} received her B.E. degree in computer science and technology from China Jiliang University in 2023. She is now a graduate student at the National University of Defense Technology, China. Her research interests include computer vision and generative models.
\end{IEEEbiography}

\vspace{-11pt}
\begin{IEEEbiography}[{\includegraphics[width=1in,height=1.20in,clip,keepaspectratio]{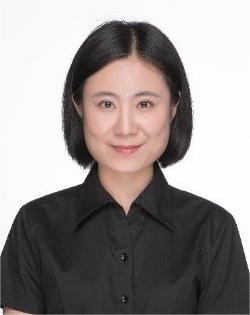}}]{Wei Chen}
is a Professor at the
School of Computer, National University of
Defense Technology. Her current research interests include computer architecture, artificial intelligence, and computer vision.
\end{IEEEbiography}

\vspace{-11pt}
\begin{IEEEbiography}[{\includegraphics[width=1in,height=1.25in,clip,keepaspectratio]{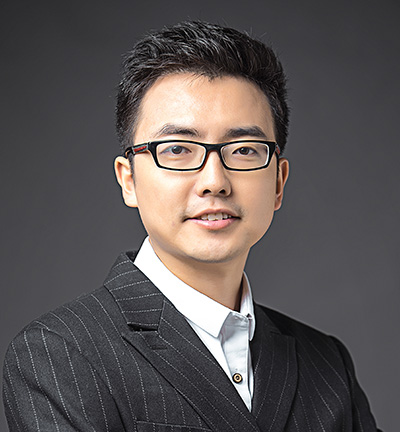}}]{Kai Xu}(Senior Member, IEEE) received the Ph.D. degree in Computer Science from National University of Defense Technology (NUDT), Changsha, China, in 2011. From 2008 to 2010, he worked as a visiting Ph.D. student with the GrUVi Laboratory, Simon Fraser University, Burnaby, BC, Canada. He is currently a Professor at the School of Computer Science, NUDT. He is also an Adjunct Professor at Simon Fraser University. His current research interests include 3D vision and embodied intelligence.
\end{IEEEbiography}

\vfill

\end{document}